\documentclass[cspaper, compsoc, 10 pt, journal]{IEEEtran}

\usepackage{graphicx}
\usepackage{amsmath} 
\usepackage[nocompress]{cite}
\usepackage{subcaption}
\usepackage{color}
\usepackage{soul}
\sethlcolor{green}
\renewcommand\hl[1]{#1}
\usepackage{algorithm}
\usepackage{algorithmic}
\usepackage{hyperref}
\hypersetup{
    colorlinks=true,
    linkcolor=blue,
    filecolor=magenta,      
    urlcolor=cyan,
}

\graphicspath{{figures/}}

\begin{document}

\title{\LARGE \bf
luvHarris: A Practical Corner Detector for Event-cameras
}

\author{Arren Glover, \textit{Member, IEEE}, Aiko Dinale, Leandro De Souza Rosa, Simeon Bamford, \\and Chiara Bartolozzi, \textit{Member, IEEE}%
\IEEEcompsocitemizethanks{\IEEEcompsocthanksitem All authors are with the Event-Driven Perception for Robotics group, Istituto Italiano di Tecnologia, Italy.\protect\\
E-mail: {\tt\small \{arren.glover, chiara.bartolozzi\}@iit.it}}
\thanks{Manuscript submitted to IEEE TPAMI April, 2021}}



\IEEEtitleabstractindextext{%
\begin{abstract}
There have been a number of corner detection methods proposed for event cameras in the last years, since event-driven computer vision has become more accessible. Current state-of-the-art have either unsatisfactory accuracy or real-time performance when considered for practical use, for example when a camera is randomly moved in an unconstrained environment. In this paper, we present yet another method to perform corner detection, dubbed look-up event-Harris (luvHarris), that employs the Harris algorithm for high accuracy but manages an improved event throughput. Our method has two major contributions, 1. a novel ``threshold ordinal event-surface'' that removes certain tuning parameters and is well suited for Harris operations, and 2. an implementation of the Harris algorithm such that the computational load \textit{per event} is minimised and computational heavy convolutions are performed only `as-fast-as-possible', i.e. only as computational resources are available. The result is a practical, real-time, and robust corner detector that runs more than $2.6\times$ the speed of current state-of-the-art; a necessity when using high-resolution event-camera in real-time. We explain the considerations taken for the approach, compare the algorithm to current state-of-the-art in terms of computational performance and detection accuracy, and discuss the validity of the proposed approach for event cameras.
\end{abstract}

\begin{IEEEkeywords}
event-driven vision, robotic-vision, event-camera, corner-detection, real-time
\end{IEEEkeywords}}

\maketitle
\IEEEpeerreviewmaketitle
\thispagestyle{empty}
\pagestyle{empty}

\IEEEraisesectionheading{\section{INTRODUCTION}}

\IEEEPARstart{C}{orner} detection is used for motion estimation and feature point identification among other machine vision tasks~\cite{luo2013survey}. In general, corners can be used as informative features that are consistently identifiable over time. For motion estimation, the presence of the two intersecting edges that define a corner disambiguates the unobservable motion in the direction parallel to a single edges orientation, i.e. it solves the aperture problem.

Several proposed corner detection methods for event-cameras investigate interesting ideas for event processing, but which we find are insufficient for actual use in a complete visual pipeline. However, event-cameras are still a promising technology for the task as they produce a low-latency, sparse visual signal and have the potential to enable high-frequency, reduced computation, visual algorithms in a wide range of applications. An event-camera achieves these advantages as it has independent, asynchronously firing pixels, rather than a global or rolling shutter. When a change in the light intensity is detected, each individual pixel outputs an \textit{event} encoding the pixel position, and the direction of the gradient of the change. As such, further development of corner detection algorithms for event-cameras is worthwhile.

Event-driven corner detectors have been used in motion estimation pipelines~\cite{Vasco2017, Vidal2018}, and a number of corner detection solutions have been proposed~\cite{Vasco2016a, Mueggler2017a, Alzugaray2018, Li2019, Manderscheid2019}. eHarris~\cite{Vasco2016a} employs the accurate Harris algorithm but is too computationally heavy for on-line use, especially with high-resolution event-cameras. FAST~\cite{Mueggler2017a} and ARC~\cite{Alzugaray2018} aim to reduce the computational requirements, but sacrifice accuracy to do so. While their event-throughput improves, they may still fail to obtain real-time performance when using the latest generation, high resolution event-cameras.

In this paper, we present yet another method for performing corner detection with event-cameras, that we dub \textit{luvHarris} for \textbf{l}ook-\textbf{u}p e\textbf{v}ent-\textbf{Harris}. Its focus is to produce both a real-time, robust and accurate corner detector based on the Harris algorithm~\cite{harris1988combined}. To do so, luvHarris decouples the event throughput from the heavy computation of the Harris algorithm. Only a small (non-corner-related) computation is performed per event, and the result is event throughput of 8.6 M events/s ($2.6 \times$ improvement over the state-of-the-art). Other detectors, while light-weight, perform a full corner detection for each and every event. We instead take advantage of OpenCV~\cite{opencv_library}, \hl{but do so with a hybrid method that sits somewhere between fully asynchronous and batch processing, but still manage to maintain an asynchronous} event-driven input and output, and discuss why this is a valid solution particularly for corner detection.

\hl{Experiments and algorithms in this paper expressly focus on corner detection only, instead of a combination of detection and tracking. Improvements to the underlying detection algorithm will always improve downstream tracking algorithms and therefore we believe evaluation of detection and tracking should be performed independently.}

This paper has two main algorithm development contributions: \textbf{1.} a novel \textit{event-surface} that is compatible with the Harris algorithm, and \textbf{2.} \hl{a look-up pipeline that enables the event-stream to be always processed asynchronously as it is decoupled from the corner detection pipeline, which is instead only processed as-fast-as-possible given the available computational power.} Additionally, this paper explains reasons for failure of other detectors, from an accuracy and event-throughput perspective, and adds to the discussion around the balance between event-driven and batch-processing that enables practical on-line vision algorithms for event-driven cameras. 

\section{BACKGROUND} \label{BACKGROUND}

The event-based adaptation of the Harris algorithm, eHarris~\cite{Vasco2016a}, became the benchmark early on as it was simple, based on a known method, and open-source code was available. The algorithm created a binary \textit{surface} indicating the occurrence of an event in the recent past, over which the Harris score was calculated. The algorithm was \textit{event-by-event} in that for each incoming event, the surface was incrementally updated, and the Harris response was computed only locally around the position on the surface at which the event occurred. The Harris response is dependent on the Eigenvalues of the image derivative, in a square patch around the event position:




\begin{itemize}
\item The partial per-pixel derivatives, $I_x = \frac{\partial}{\partial x}$ and $I_y = \frac{\partial}{\partial y}$, are calculated using the Sobel operator.

\item The mean-square derivatives of the patch, $G_{xx} = I_x \times I_x$, $G_{xy} = I_x\times I_y$, and $G_{yy} = I_y\times I_y$, are calculated by element-wise multiplication through a box filter.
    
\item The Harris response is related to the Eigenvalues of the matrix $M=\begin{bmatrix} G_{xx} & G_{xy} \\ G_{xy} & G_{yy}\end{bmatrix}$, and is calculated as $R = det(M) - 0.04 \cdot tr(M)^2$, where $det()$ and $tr()$ are the determinant and trace, respectively.
\end{itemize}
An event is classified as a \textit{corner-event} if the Harris response is above a threshold $T_R$.


The accuracy of corner classification using the eHarris method was comparable to a `frame-based' ground-truth, however the processing speed was sub-par~\cite{Mueggler2017a, Alzugaray2018} for high-rate event-streams. eHarris was developed for the first generation DVS~\cite{Lichtsteiner2008} ($128 \times 128$ resolution), but the reported real-time operation for a data stream of 160k events/s cannot handle higher resolution sensors (e.g. the qVGA ATIS~\cite{Posch2008} or DAVIS~\cite{Brandli2014}) that were becoming the standard. Nowadays HD event-cameras are also available. 

The subsequent FAST~\cite{Mueggler2017a} and ARC~\cite{Alzugaray2018} algorithms were proposed to provide a solution for higher-resolution event-cameras, and do so by moving away from traditional image processing techniques, towards techniques that are highly compatible with the asynchronous, fine temporal resolution of event-cameras.
Indeed, these techniques reported throughput an order of magnitude higher than eHarris, greatly increasing the conditions in which real-time processing is maintained. FAST reported comparable corner detection accuracy to eHarris~\cite{Mueggler2017a}, \hl{while ARC classified more events as corners (including more false positives) which were then filtered with a downstream, off-line, corner tracking algorithm to produce robust corner tracks}~\cite{Alzugaray2018}.

The FAST and ARC algorithms are \textit{arc-based} detection methods that identify corners by classifying the spatio-temporal pattern on a \hl{surface of active events (SAE)}:

\begin{equation}
SAE : (x, y) \mapsto t
\end{equation}
\noindent where $t$ is the event timestamp. A continuous arc of recent timestamps which covers an approximate 90 degree angle results in a classified corner, while broken arcs, or arcs that cover an angle close to 180 degrees can be rejected.

\begin{figure}
	\centering
	\subcaptionbox{\label{fig:arcprob_1}}{
	\includegraphics[width=0.54\linewidth]{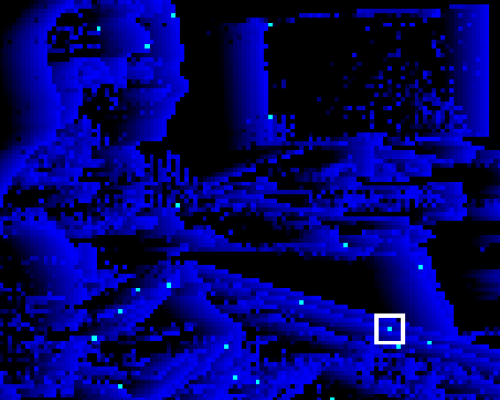}
	\includegraphics[width=0.45\linewidth]{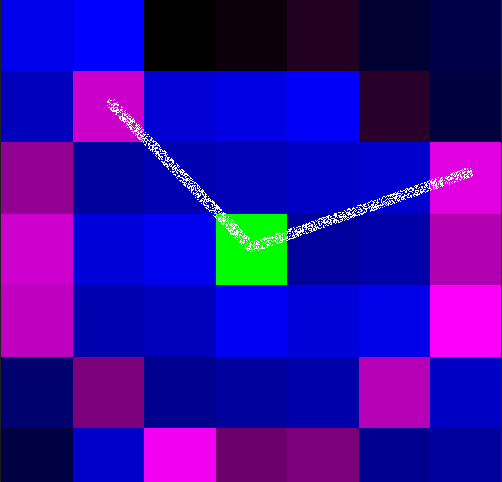}}\\
	\subcaptionbox{\label{fig:arcprob_2}}{
	\includegraphics[width=0.53\linewidth]{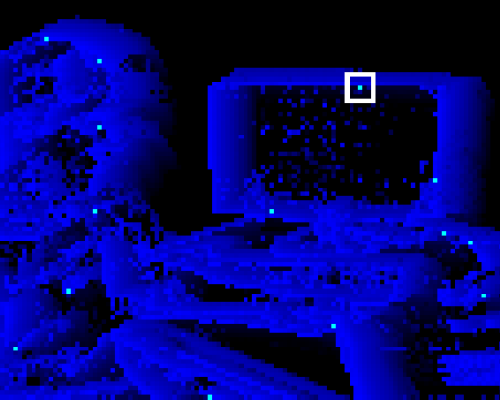}
	\includegraphics[width=0.45\linewidth]{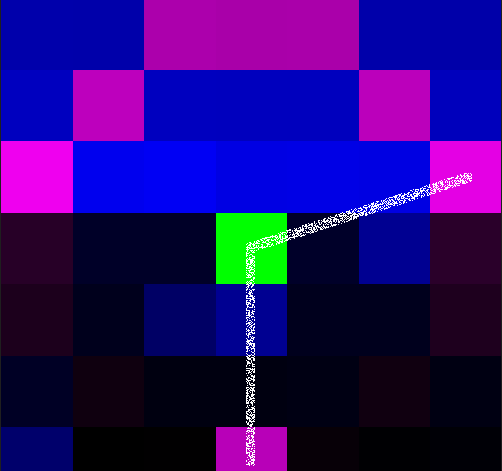}}\\
	\subcaptionbox{\label{fig:arcprob_3}}{
	\includegraphics[width=0.55\linewidth]{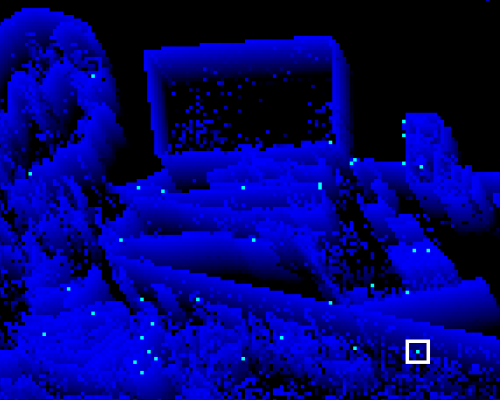}
	\includegraphics[width=0.44\linewidth]{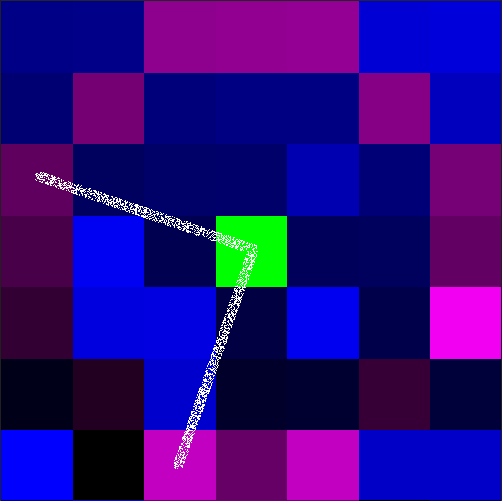}}
	\caption{\hl{Examples of false positive ARC corner detections: outlined events (left) are wrongly selected as corners but given their position in the image are clearly not corners. 
	The brightness of the pixel in the SAE (right) corresponds to the time value of the surface (black being older events) and the white lines correspond to the final arc selection. Purple pixels indicate only values used for an arc-based corner detector. (a) occurs as two edges are close together, (b) occurs due to a single random event off the edge, and (c) occurs in clutter in which an arc is found by chance. Using all points in the local time surface (not only purple) gives more information to determine there is no corner.}}
	\label{fig:example2}
\end{figure}

\subsection{Issues with arc-based detection}\label{sec:arcissues}

\hl{Arc-based detections use only a circle of events around the corner candidate to determine the corner from the spatio-temporal pattern. They use less information (less pixels) to determine corner patterns, compared to Harris, and are therefore more susceptible to noise in those pixels.} Fig.~\ref{fig:example2} \hl{shows several examples of incorrect corner detections of an arc-based method, which, when considering the entire patch, are obviously not patterns of data that correspond to corners.

Typically it is an unexpected SAE value (i.e. timestamp) that occurs in-front of the edge of the current event, that causes an incorrect positive classification:}
\begin{itemize}
    \item \hl{An object that has two straight parallel edges close to each other can create an interference, in which the pixels of the first edge are counted within the arc}, see Fig.~\ref{fig:arcprob_1}. In this case the active pixel can be incorrectly classified as a 270 degree corner.
    \item \hl{Jitter-like motion that causes the edge to move back on itself multiple times can produce an SAE that has the characteristics similar to that of two parallel edges as above.}
    \item \hl{In reality, edge-gradients are non-discrete, resulting in a single edge producing multiple events at the same location over a short period of time. These \textit{secondary} events occur after the leading edge and result in an SAE similar to} Fig.~\ref{fig:arcprob_1}. Filters can be used to remove \textit{secondary} events~\cite{Alzugaray2018} \hl{but should be tuned to the temporal dynamics of the application, which isn't always a robust solution.}
    \item \hl{Noise can manifest in the event-stream as randomly firing pixels. A random event nearby a moving edge will trigger a corner detection as the pixel becomes one of the end-points of the arc}, see Fig.~\ref{fig:arcprob_2}. Filters can also be used to reduce the frequency of random events.
    \item \hl{Clutter, or highly textured patches can also satisfy the corner detection logic by chance}, see Fig.~\ref{fig:arcprob_3}, despite no clear corner pattern in the local patch.
\end{itemize}

\hl{Some of the above sources of incorrect detections are directly addressed in the ARC algorithm by applying a pre-processing step to the event stream}~\cite{Alzugaray2018}. The pre-processing consists of an event filter that behaves as an \textit{artificial refractory period}. The authors suggest a 50 ms artificial refractory period, however such a large filter period also removed large portions of useful signal, in which corners were not detected. In this case, the maximum firing rate corresponds to a 20 Hz repeating signal, in which case a RGB camera may be preferable. Instead, a 5 ms refractory period seemed suitable for typical signals from a hand-held, or robot-mounted, camera. However, more false positives were observed with the lower value. The requirement of such a pre-processing imposes restrictions on the use-cases of the ARC algorithm.

The FAST algorithm also addresses some of the sources of incorrect detections as it excludes 270 degree corners in the classification stage~\cite{Mueggler2017a}. FAST does not require a pre-processing stage, but therefore also cannot detect certain corner patterns (i.e. those produced by 270 degree corners) that are true corners.

\subsection{Considerations for Applying the Harris algorithm to event-data}\label{sec:considerations}

The dense computation of spatial derivatives required by Harris has the downside of a higher processing requirement. Reducing the load of the Harris pipeline has been suggested by using FAST as a candidate detector, and applying Harris only to positive candidates~\cite{Li2019}. We instead take a different approach: to enable a real-time Harris-based detector by exploiting the principles of the algorithm.

Original Harris corner detection is applied over an image, in which pixel values are typically bound between 0 and 255. The magnitude of a corner score, $R$, is proportional to the difference, $d$, between the high and low values in a patch. To fix a classification threshold, $T_R$, $d$ values associated to corners must be consistent over the entire image and over time. In the event domain, an SAE is an obvious choice to represent amalgamated data, however it does not produce a consistent $d$. High values will correspond to the current clock-tick, while low values will be equal to the time in the past when the pixel last fired, which could be on the order of milliseconds, or minutes; it is therefore difficult to fix a value for $T_R$. For this reason, the binary image used by original eHarris~\cite{Vasco2016a} was not a simplistic, naive implementation, but rather a calculated decision to for consistent $d$ values. However, the particular solution was dependent on a parameter that defined the typical texture found in the scene \hl{(i.e. using only the most recent e.g. 1000 pixels).}

Secondly, as a corner is defined by relative values between neighbouring pixels, computation can be saved by re-using the convolution results between neighbouring convolutions. As such, the convolution computation, $P_c$, required for two neighbouring pixels is not simply $P_{c1} + P_{c2}$, but $P_{c1} + (1-o)P_{c2}$, where $o$ is the proportion of overlap between the two regions. I.e. computing all spatial convolutions simultaneously across an image is faster than computing the convolution independently for each pixel. Performing Harris event-by-event throws away the intermediate calculations that can instead be re-used for neighbouring convolutions. 

An interesting alternative approach to apply this principle to event-cameras, is to calculate gradient convolutions \textit{incrementally} as in~\cite{Scheerlinck}, \hl{or a training a neural network to estimate gradients from event volumes}~\cite{chiberre2021detecting}. However, \cite{Scheerlinck} uses a fixed decay parameter limiting use for variance in scene dynamics, and \cite{chiberre2021detecting} \hl{uses a fixed number of events in an input volume, limiting use in scenes with different amounts of texture. Both methods also involve a batch operation sequentially in the event pipeline removing the asynchronous, and low-latency properties of the system. However, an interesting insight from both studies is the implication that corner detection is solved using a Harris algorithm, as long as the input is well formed. Such an insight highlights the importance of creating the underlying representation correctly.}

Considering the above, luvHarris has been designed with the following principles:
\begin{itemize}
\item A full patch of the surface gives more accurate corner classification than an arc, therefore the Harris method is chosen. 
\item The surface over which Harris is applied needs to have consistent maximum and minimum values to enable a fixed classification threshold to be used and, for robustness, should avoid arbitrary parameters (i.e. a temporal window).
\item A corner is a spatial pattern, therefore spatial convolutions (or other pattern matching) is required. Redundant processing can be avoided by re-using convolutions results for corner classification of neighbouring pixels.
\end{itemize}
\newpage

\section{Look-up event-Harris} \label{section:two}

The proposed luvHarris is comprised of two parts:
\begin{itemize}
\item An \textit{event-by-event} update of a threshold-ordinal surface (TOS). The TOS is designed to accumulate visual data so it is compatible with the Harris corner detection algorithm, and does so without the need for an arbitrary temporal parameter.
\item An \textit{as-fast-as-possible} computation of a Harris-score look-up-table, $L$: when an instance of Harris computation is finished, a new one begins, independently from the number of events that need to be processed.
\end{itemize}

\begin{figure}[h]
	\centering
	\includegraphics[width=1.0\linewidth]{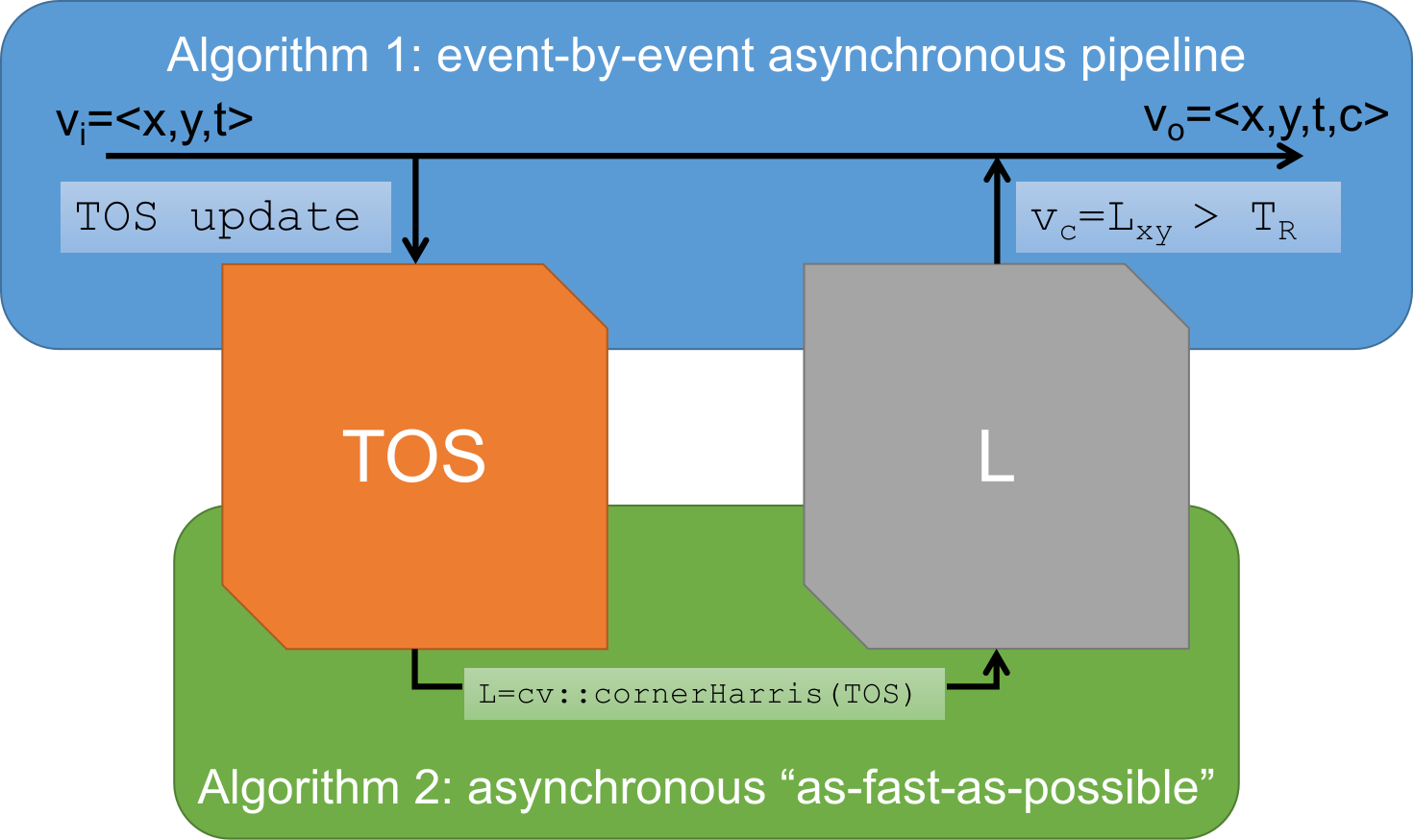}
	\caption{The flow of data through the system. \hl{Algorithm 1 is completely asynchronous and performed event-by-event, assigning corner labels to each event. Algorithm 2 is decoupled from the processing schedule of Algorithm 1 and provides an up-to-date 2D look-up table of corner scores for each pixel, as often as possible based on the hardware. The algorithms can be computed simultaneously if multiple cores are available. If only a single core is available, luvHarris can be run switching between algorithms sequentially; in this case Algorithm 1 must be computed for all `already-produced' but unprocessed events.}}
	\label{fig:blocks}
\end{figure}

Our goal is to implement an on-line system in which events are streamed live from a camera, and the processing of corners must be real-time, for as high-as-possible event-rate. Events are streamed into the processing module, and events tagged as corners (and not-corners) are streamed out; hence the asynchronous nature of events is fully  maintained in luvHarris. An overview of the main algorithm components are illustrated in Fig.~\ref{fig:blocks}, and each block is explained below.

\subsection{Event-by-event computation: Threshold-ordinal Surface}

\begin{algorithm}[b]
\caption{Event-by-event computation ($q_1$)}
\label{alg:TOSalg}
\begin{algorithmic}
\REQUIRE $v = \langle x,y,t \rangle$, $TOS$
\STATE \verb|for| $x = v_{x}-k_{TOS} : v_{x}+k_{TOS}$
\STATE \quad\verb|for| $y = v_{y}-k_{TOS} : v_{y}+k_{TOS}$
\STATE \qquad$TOS_{xy} \gets TOS_{xy} - 1$
\STATE \qquad\verb|if |$TOS_{xy} < 255 - T_{TOS}$
\STATE \qquad\quad$TOS_{xy} \gets 0$
\STATE $TOS_{v_xv_y} \gets 255$
\STATE $v_c \gets L_{xy} > T_R$
\end{algorithmic}
\end{algorithm}

\hl{The event-by-event computation is fully asynchronous and is comprised of two parts: 1. the update to the TOS, and 2. the assignment of the corner score.}

\begin{figure}
	\centering
	\subcaptionbox{}{
	\includegraphics[width=0.55\linewidth]{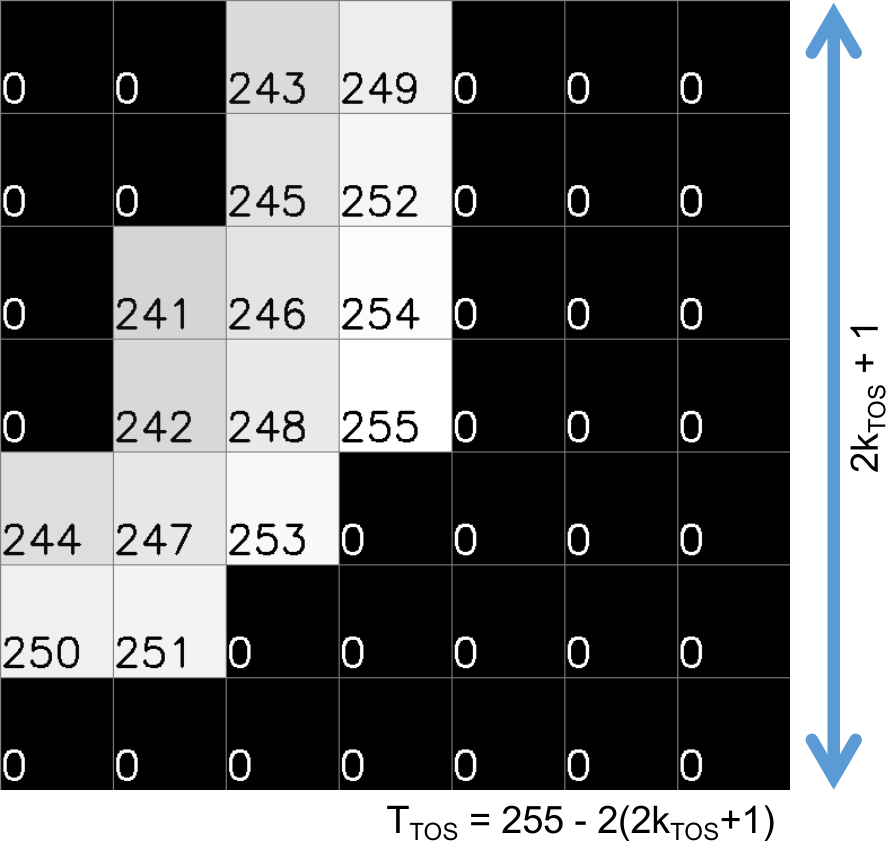}}\\
	\subcaptionbox{}{
	\includegraphics[width=0.48\linewidth]{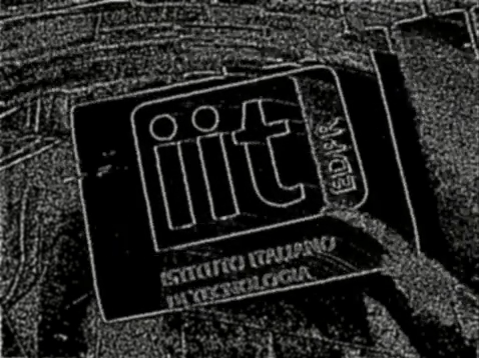}}
	\subcaptionbox{}{
	\includegraphics[width=0.48\linewidth]{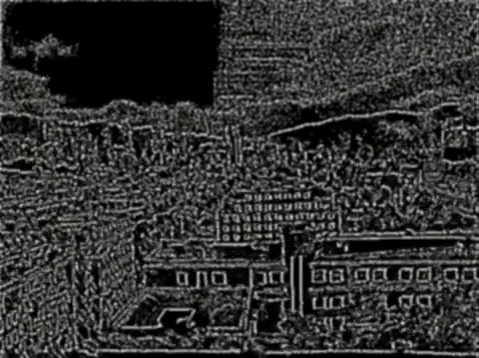}}
	\caption{Examples of the threshold-ordinal surface \hl{(a) update patch, and (b-c) the full surface visualised as an image. (a) is produced with $k_{TOS}=3$ leading to $T_{TOS}=241$, below which all values are set to 0. The brightness of each pixel represents the value in the TOS. (b) and (c) show that strong edges and corners are present in the visual signal, while blank regions are either zero (black), or filled with random noise, neither of which produce a strong corner response.}}
	\label{fig:exampleTOS}
\end{figure}

The TOS, visualised in Fig.~\ref{fig:exampleTOS}, provides a coherent and bound spatial representation of the asynchronous events, partially maintaining the information about their temporal order and attempts to capture the most up-to-date position of edges in the scene. The TOS is defined as:

\begin{equation}
TOS : (x, y) \in 0 \cup [(255-T_{TOS}) \rightarrow 255]
\end{equation} 
and for any input event $v_i = \langle x,y,t \rangle$, the full processing to update the TOS and assign the corner classification $v_c$ is defined in Algorithm~\ref{alg:TOSalg}, where $k_{TOS}$ is a local region half-size, $T_{TOS}$ is a threshold, $T_R$ is the Harris score threshold, and $L$ is the \hl{2D} look-up table explained \hl{in Section}~\ref{sec:L}.

\hl{The parameter $k_{TOS}$ defines the local region size and the same value is used for the Harris detection region. The appropriate selection of $k_{TOS}$ comes from the relative size of objects and the camera resolution.}

\hl{The parameter $T_{TOS}$ defines the approximate desired number of non-zero TOS values in any region of size $2k_{TOS}+1$. As the TOS value is set to 255 for new events, and new neighbouring events subtract 1 from the entire region, once a TOS values reaches $255-k_{TOS}$ setting it to 0 achieves this desired number (under ideal conditions). The appropriate selection of $T_{TOS}$ for corner detection is made to form an edge of 2 pixels thick and $T_{TOS} = 2 \cdot (2k_{TOS}+1)$. For a typical patch size of $7\times7$ ($k_{TOS}=3$), $T_{TOS}$ will take the value of 14, which corresponds to a line of 7 pixels, 2 pixels thick, for a perfect clean edge passing through the region.}

The TOS is somewhat similar to the Speed Invariant Time Surface~\cite{Manderscheid2019} (SITS), in which it was concluded that the ordinal method (the surface value corresponds to the order of event arrival) promoted speed invariant representations suitable for corner detection. To apply the considerations presented in Section~\ref{sec:considerations}, the following differences are implemented:
\begin{itemize}
    \item \hl{The TOS has an unused range between 0 and $255-T_{TOS}$ which increases the signal-to-noise ratio of gradients for the Harris algorithm. Instead, the SITS has a range $0 \rightarrow (2r+1)^2$ with $r \equiv k_{TOS}$. In this case, gradient slopes can be of the same magnitude as noise.}
    \item \hl{Under good conditions the value of the TOS is forced to return to 0 after an edge has passed, due to the threshold $T_{TOS}$. The SITS is more likely to have some unknown value after an edge has passed. Therefore the SITS can result in inconsistent gradient calculations depending on the history of the SITS at that location.}
    \item \hl{Concerning the specifics of implementation: the TOS reduces the values of all values in the local region to calculate the true ordinal value, while the SITS only reduces values which are larger than that of the current active pixel location.}
\end{itemize}

\subsection{As-fast-as-possible computation: Harris Look-up}\label{sec:L}

The TOS can be used at any given point in time as an `image', with values between 0 and 255, on which to perform the Harris calculation using the \texttt{cv::cornerHarris} function from OpenCV~\cite{opencv_library}. \hl{The output of \texttt{cv::cornerHarris} is a 2D array which is directly used to populate the 2D look-up-table, $L$, such that the value at each entry is the Harris score.}

\begin{algorithm}[b]
\caption{As-fast-as-possible computation ($q_2$)}
\label{alg:harris}
\begin{algorithmic}
\STATE $L \gets$ \verb|cv::cornerHarris(TOS)| 
\end{algorithmic}
\end{algorithm}

\hl{Instead of applying a threshold to the output of \texttt{cv::cornerHarris} and searching the array for pixels that satisfy the threshold, in luvHarris $L$ is directly populated from this output. $L$ is queried sparsely and asynchronously by the event-stream and the threshold $T_R$ is therefore also applied sparsely.} If $L_{xy} > T_R$ then the event is tagged as a corner.

One key aspect of luvHarris is that the data flow of events is completely decoupled from the generation of $L$: the TOS is updated asynchronously event-by-event, and $L$ is computed over the most up-to-date TOS as frequently as computation allows. As $L$ cannot be feasibly generated for each single event, a single instance of $L$ is used for multiple events. The value $t_{err} = L_t - v_t$, describes the temporal error between the time of an event, $v_t$, and the time in which $L$ was generated, $L_t$; the likelihood of event mis-classification increases with $t_{err}$. However, the relationship is not linear, and there exists a reasonable period in which $t_{err}$ results in little to zero change in the Harris score $L_{xy}$. If a corner completely passes through a patch (i.e. defined by $k_{TOS}$) between updates of $L$, that corner will be `missed' by the algorithm, however due to the nature of Harris the algorithm has several pixels of tolerance.

\subsection{Throughput Limitations}

Real-time processing is achieved despite a variable number of input events, as the processing for each event (TOS) is decoupled from heavy algorithm processing (Harris). The total processing, $P$, of an event-based algorithm is: 

\begin{equation} \label{eq:P}
P = q_1V + q_2W
\end{equation}
where $q_1$ is the number of computations required \textit{per-event}, $V$ is the total number of events, and $q_2$ is the number of computations performed for, $W$, \textit{non-event-based} processes. If $P > P_{max}$, where $P_{max}$ is the maximum number of operations capable by the CPU in the period in question, then the algorithm does not operate in real-time.

All previous state-of-the-art corner detection methods perform all corner detection calculations encapsulated by $q_1$, and there are no non-event-based computations, i.e. $q_2 = 0$. Therefore the total computational requirements are directly proportional to the number of events that are generated by the camera. In general, fully event-by-event algorithms suffer a reduced event throughput as the algorithm complexity increases (given a fixed $P_{max}$). While FAST and ARC are computationally light algorithms, their maximum event throughput (2.0M, and 3.3M events/s respectively \hl{after filtering}) are less than next-generation event-cameras (typically over 4M events/s \hl{after filtering}). 

The general contribution of luvHarris to event-driven algorithms is to shift processing from $q_1$ to $q_2$, to minimise the computational requirements dependent on the event-rate without completely removing $q_1$ (which would result in fully batch-based computation). For luvHarris, $q_1$ corresponds to updating the TOS and a look-up of a single pixel value in $L$, for each event, while $q_2$ encapsulates the Harris computation, which is independent from the number of events that must be processed.

luvHarris also lends itself well to parallel operation: as Algorithm~\ref{alg:TOSalg} and Algorithm~\ref{alg:harris} are completely decoupled, they can be simultaneously processed on separate CPU cores, as illustrated in Fig.~\ref{fig:blocks}. However, on a single-core CPU real-time operation can still be achieved by running the Algorithms sequentially. Algorithm~\ref{alg:TOSalg} must process all events already produced and queued by the camera, after which Algorithm~\ref{alg:harris} is performed once before switching back to Algorithm~\ref{alg:TOSalg}. As the event-rate increases, the number of events that must be processed in a single `batch' increases and corner misclassification can occur as more events are associated with a single $L$, however the overall algorithm latency is minimised\footnote{under the assumption the CPU is powerful enough to process Algorithm~\ref{alg:TOSalg} for each and every event, \hl{otherwise luvHarris will not achieve} real-time performance.}.

\section{EXPERIMENTS AND RESULTS} \label{section:exp}

The luvHarris algorithm is compared to eHarris, FAST and ARC algorithms in terms of real-time operation and corner detection accuracy. The \texttt{shapes\_6dof, poster\_6dof, boxes\_6dof}, and \texttt{dynamic\_6dof} datasets~\cite{EDdatasets} are used, as in~\cite{Mueggler2017a, Alzugaray2018}\footnote{In our case we used only the 6-DoF datasets as the literature has shown them to be the most challenging}. The \texttt{boxes\_6dof} and \texttt{poster\_6dof} datasets were trimmed due to technical limitations loading the full dataset into memory.  Additionally we perform on-line experiments with an ATIS ($480 \times 360$ resolution) camera running all algorithms live simultaneously, \hl{with a raw event-rate of over 10M events/s, which becomes approximately 4M events/s if the event-filter is applied}.

Computation is performed on an Intel Core i7-9750H CPU @ 2.60GHz $\times$ 12. $k_{TOS}$ was set to 3, with an approximate region size equal to FAST and ARC, which use two rings of radius 3 and 4 as in their respective publications. \hl{eHarris was implemented using the spatially-adaptive method proposed in}~\cite{Mueggler2017a} \hl{using an identical region size and choosing $T_{TOS}$ recent events, such that the underlying representation was somewhat similar to luvHarris. The openCV \texttt{cv::cornerHarris} method was computed for the local patch event-by-event.} 

\hl{Algorithms are implemented ignoring the event polarity and considering a single event-stream. Such an implementation varies from some original implementations, in which polarities are computed separately. Such a choice may result in slight variation of results, however the relevance of results is not expected to be impacted, and all algorithms are computed similarly such that the comparison between algorithms should remain valid.}

 The full results are also shown as a  \href{https://zenodo.org/record/4739290}{video}\footnote{https://zenodo.org/record/4739290}.
 
 A filtered set of each dataset was also produced that removed events that occur consecutively at the same pixel location in a short time-window (i.e. artificial refractory period), or that are not correlated to their neighbours (i.e. salt-and-pepper noise) (as explained in~\cite{Alzugaray2018}). Results indicate if the \textit{filtered} or \textit{non-filtered} data was used for the experiment. \hl{We consider that the event-filter reduces the event-rate of the dataset, rather than improving the algorithm throughput. All algorithm comparisons are performed equally for all algorithms regarding the use of the filters, therefore while luvHarris does not explicitly need filtering, when appropriate it is also tested using the filtered datasets to give a valid comparison.}

\subsection{Event Throughput}

Table.~\ref{tab:event_rates} shows the maximum event-throughput measured for all algorithms by operating the algorithms at their limits. We verify that our implementations of the state-of-the-art algorithms are computationally on-par with available implementations as the values agree with those presented in~\cite{Mueggler2017a, Alzugaray2018} accounting for some variation in the exact implementation and the hardware used. Our metric considers the event-throughput of the algorithms themselves, independently of any filtering steps, which could be used on all algorithms identically, or reading events from communication pipelines. The luvHarris method has an approximate $2.6 \times$ speed improvement over the next best ARC algorithm. Increasing the $k_{TOS}$ parameter will decrease the event-throughput of the luvHarris algorithm.

\begin{table}
    \centering
    \begin{tabular}{r|c|c}
         & Measured (M event/s) & Reported (M events/s)\\
         \hline
         eHarris & 0.16 & 0.14 \\ 
         FAST & 1.98 & 1.67\\ 
         ARC & 3.27 & 7.52*\\ 
         luvHarris & \textbf{8.59} & -
    \end{tabular}
    \caption{Maximum event throughput measured compared to that reported in the literature~\cite{Alzugaray2018}. Inconsistencies arise from the exact method of measurement and hardware used, however we report general agreement. \hl{*ARC throughput reported in the literature considered the throughput of the event-filter and detection algorithm together. Instead we consider the filter as a reduction on the dataset event-rate. Therefore we measure the maximum throughput of the detection component of the algorithm, independently if the filter is used or not. The discrepancy comes from the way the measurement is expressed rather than an algorithm difference.}}
    \label{tab:event_rates}
\end{table}

Instantaneous delay should arise when, at any point in time, the event-rate of the camera exceeds the maximum event-throughput listed in Table~\ref{tab:event_rates}. Delay was measured during operation by calculating the difference between the timestamp of the \hl{most recent event in a current event packet}, and the amount of time passed since beginning the experiment. Both the experiments that use datasets and those with a real camera were performed identically, in an on-line fashion: by directly connecting the camera or streaming the datasets with precise timing to the corner detection modules. The major difference to typical off-line processing is that in our experiments events cannot be processed if they have not yet been produced. The algorithms cannot compensate high-throughput periods by also processing low-throughput periods and taking the average.

\begin{figure}
	\centering
	\subcaptionbox{\texttt{boxes\_6dof}}{
	\includegraphics[width=0.85\linewidth]{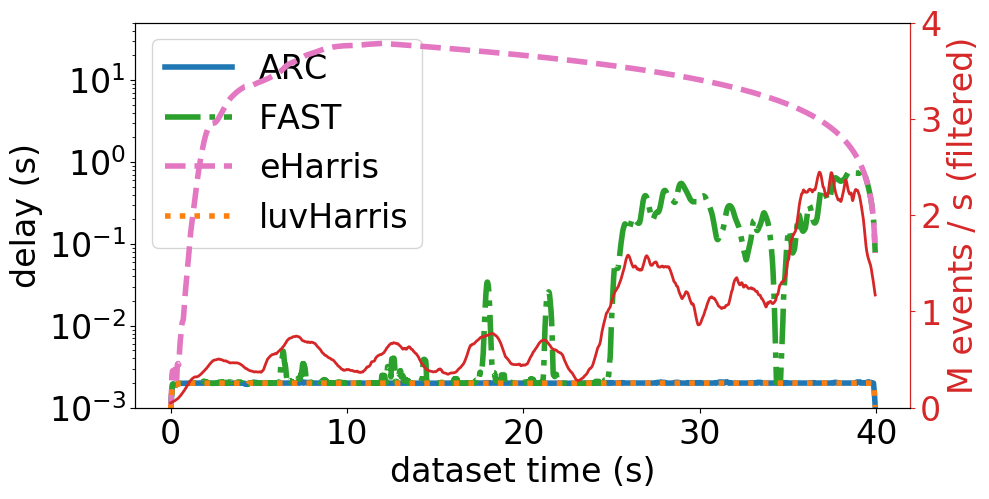}}\\
	\subcaptionbox{\texttt{dynamic\_6dof}}{
	\includegraphics[width=0.85\linewidth]{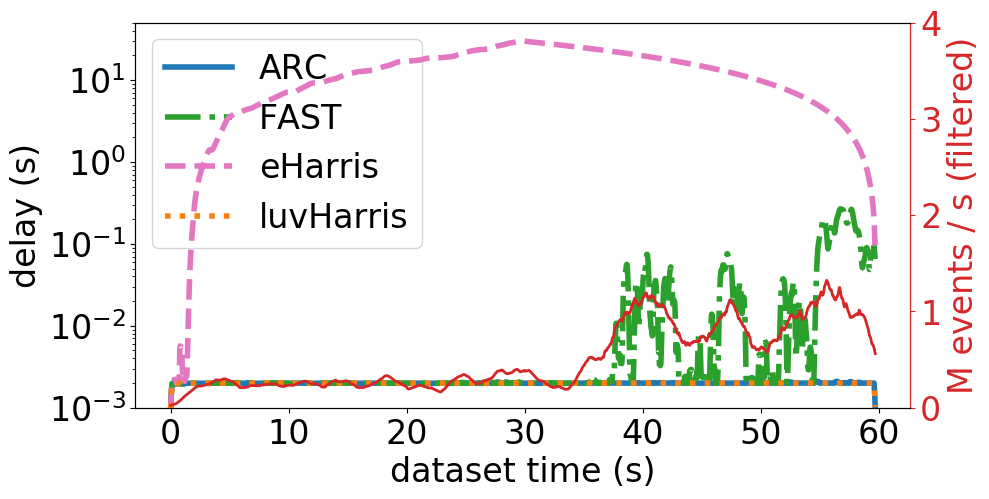}}\\
	\subcaptionbox{\texttt{poster\_6dof}\label{fig:delay_poster}}{
	\includegraphics[width=0.85\linewidth]{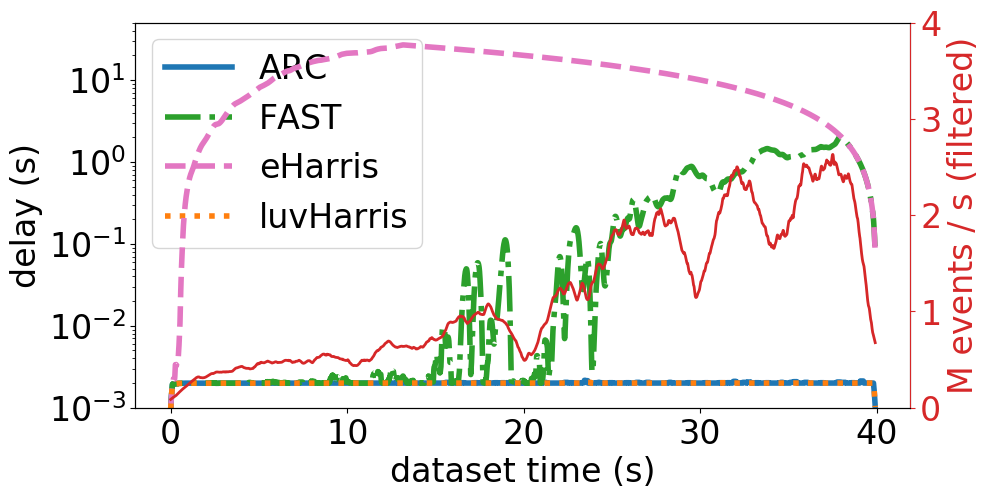}}\\
	\subcaptionbox{\texttt{shapes\_6dof}}{
	\includegraphics[width=0.85\linewidth]{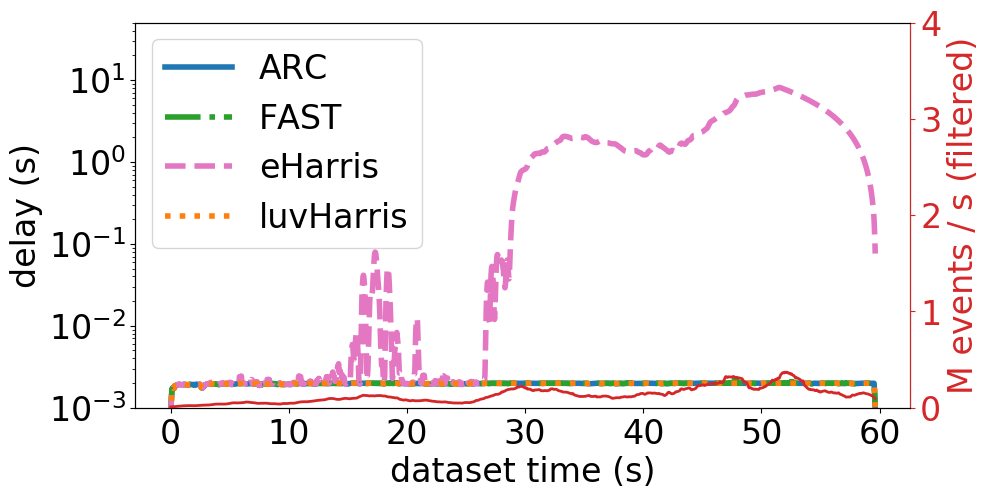}}\\
	\subcaptionbox{live with ATIS gen3\label{fig:delay_live}}{
	\includegraphics[width=0.85\linewidth]{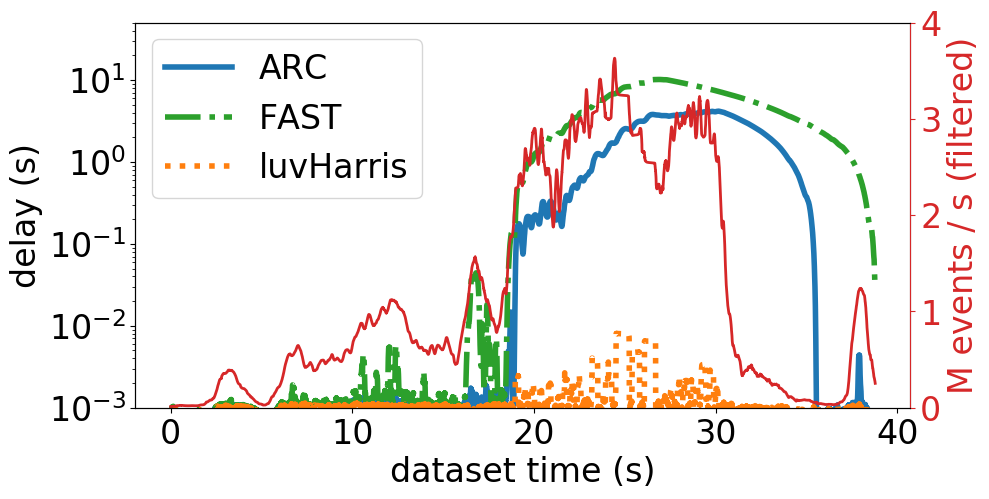}}
	\caption{Computed algorithm delay when streaming the noise-filtered datasets into the algorithm module. Any value above zero implies the algorithm is not running real-time. eHarris is not practically possible to run on-line with the Gen3 ATIS and is not present in (e). \hl{In (a)-(d) both luvHarris and ARC overlap and exist on the same line (best viewed in colour).} }
	\label{fig:main_delay}
\end{figure}

Fig.~\ref{fig:main_delay} quantitatively analyses the real-time performance of all algorithms for each dataset. The eHarris implementation has the highest delay accumulation, and is non-real-time for all datasets. FAST achieves real-time for some datasets, but not for high-texture datasets during periods of fast motion, e.g. the second half of the \texttt{poster\_6dof}, as indicated by a non-zero delay. Both ARC and luvHarris managed to maintain real-time for all instances in all datasets. The delay results achieved completely agree with the real-time assessment in~\cite{yilmaz2021evaluation} and the expected throughput reported in Table~\ref{tab:event_rates}.

The ARC algorithm experienced some algorithm delay when the on-line experiment reached approximately 3M events/s (see Fig.~\ref{fig:delay_live}). The value is slightly lower than reported in Table~\ref{tab:event_rates} \hl{as the measured delay additionally incorporates overheads from the refractory period filter, communications and reading of events. The luvHarris algorithm also experienced small instantaneous delays, which can arise from thread scheduling policies in a non-real-time kernel, and synchronisation between reading events simultaneously for all processing modules as well as the points mentioned for the ARC algorithm.} The delay of luvHarris is several orders of magnitude smaller than the other algorithms, given the log scale of the delay axis.

\subsection{Corner Accuracy}

\begin{figure*}
	\centering
	\subcaptionbox{The image is the output of e2vid~\cite{e2vid} network. \hl{Blue pixels indicate events associated with the particular frame for assigning ground-truth corner scores, with green pixels indicating positive corner labels.}\label{fig:main_gt}}{
	\includegraphics[width=0.23\linewidth]{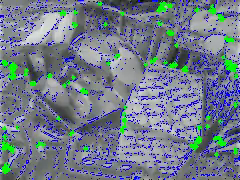}
	\includegraphics[width=0.23\linewidth]{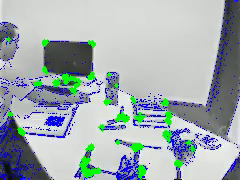}
	\includegraphics[width=0.23\linewidth]{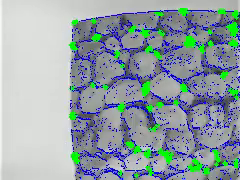}
	\includegraphics[width=0.23\linewidth]{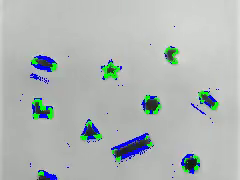}}\\
	
	\subcaptionbox{Precision-recall for each of the above datasets \textit{without} applying filtering.\label{fig:main_pr}}{
	\includegraphics[width=0.23\linewidth]{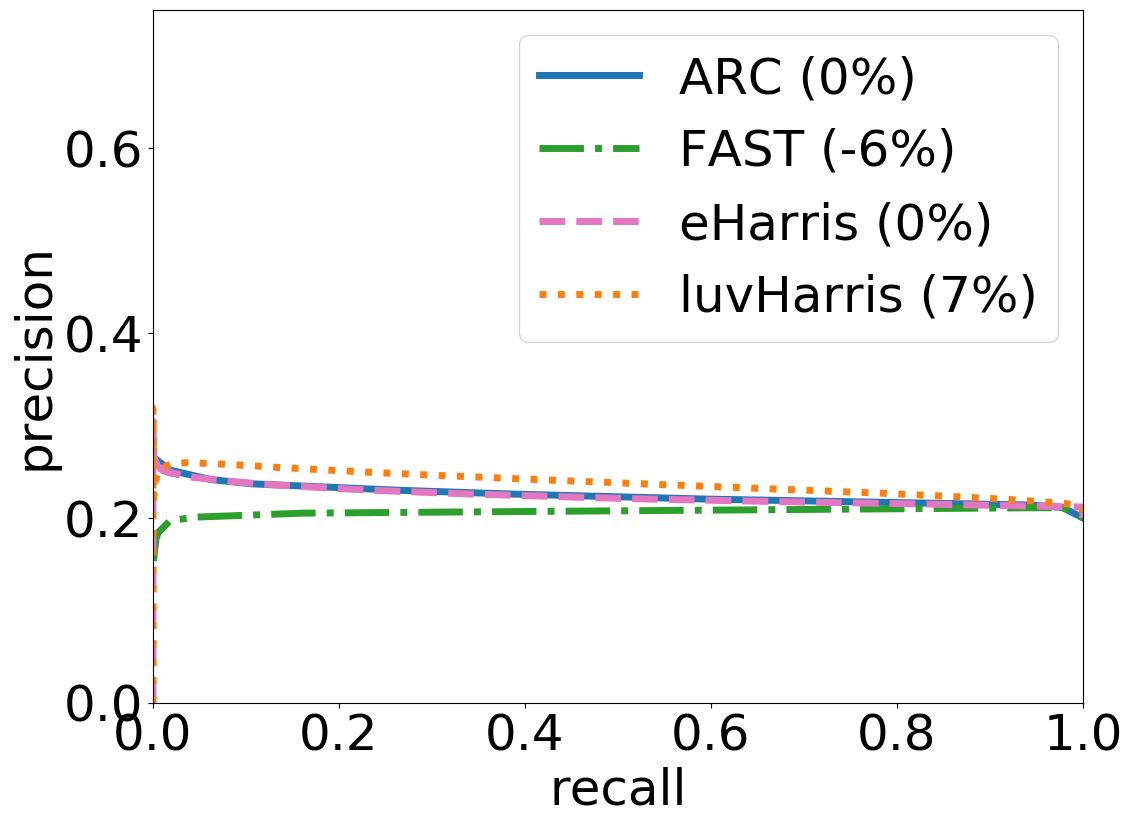}
	\includegraphics[width=0.23\linewidth]{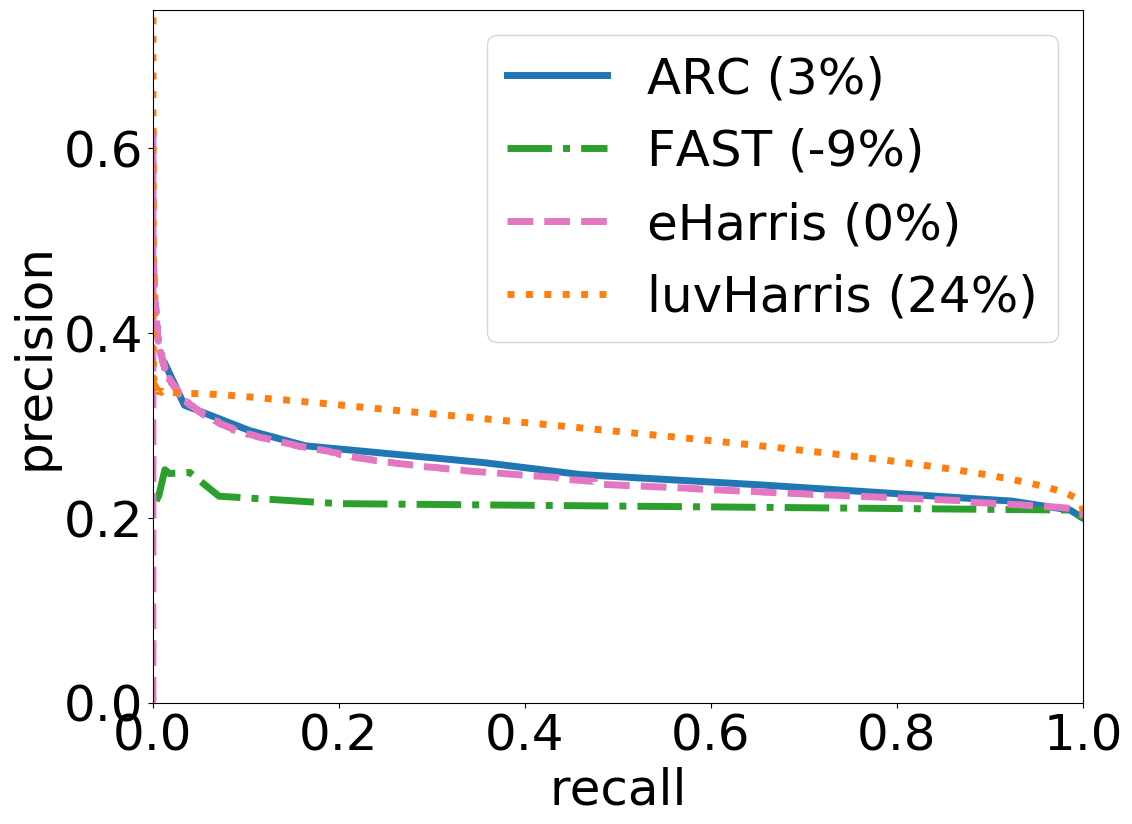}
	\includegraphics[width=0.23\linewidth]{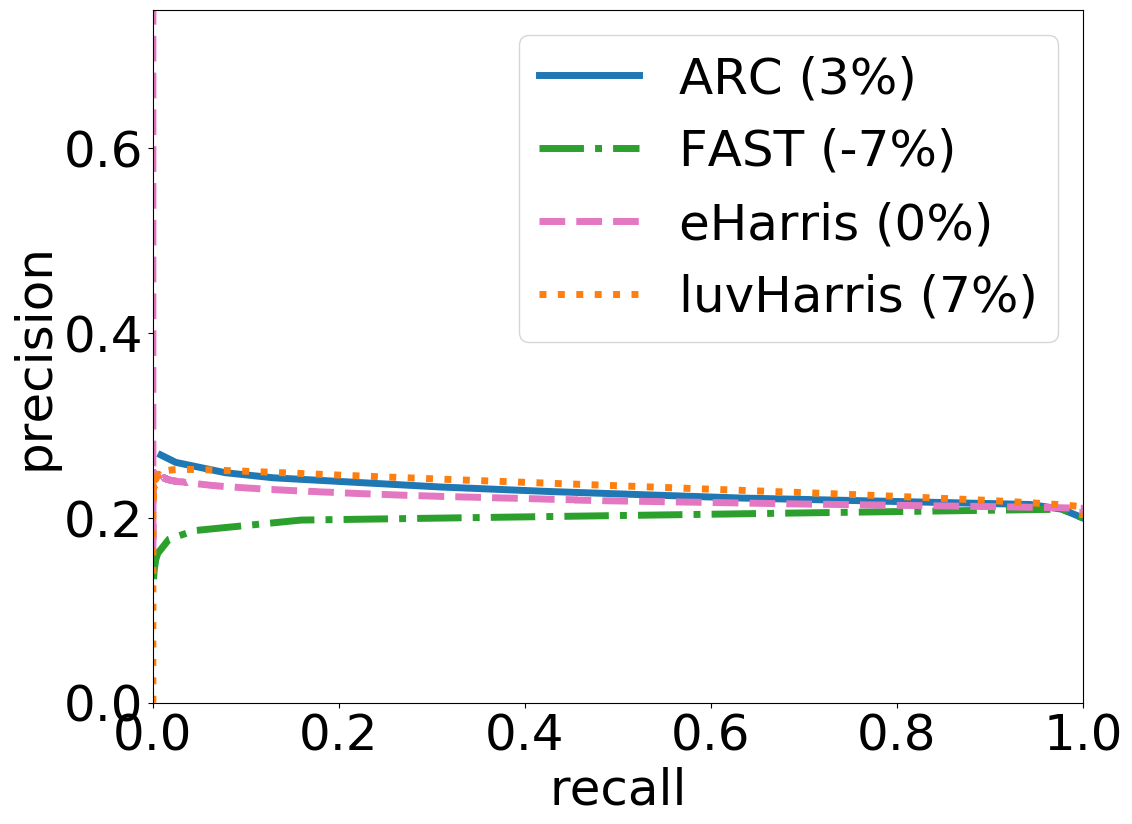}
	\includegraphics[width=0.23\linewidth]{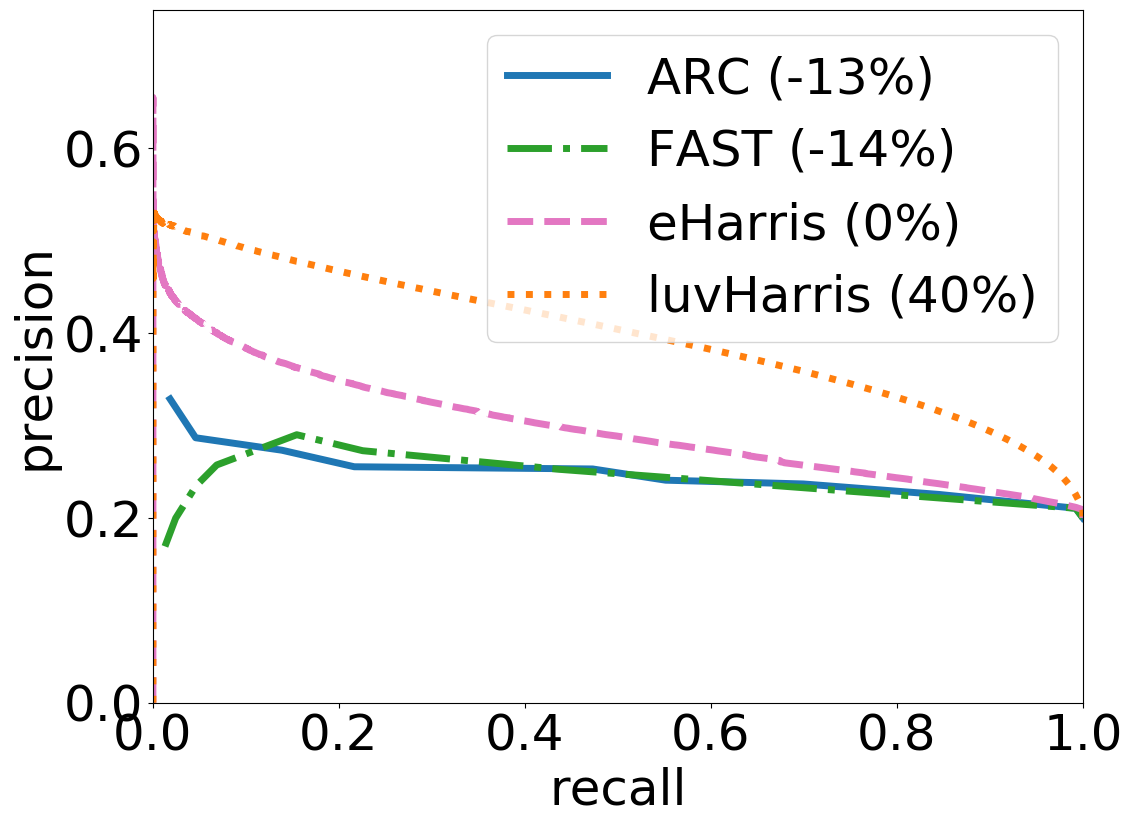}}\\
	
	\subcaptionbox{Precision-recall for each of the above datasets \textit{with} applying the filtering as in~\cite{Alzugaray2018}.\label{fig:main_prf}}{
	\includegraphics[width=0.23\linewidth]{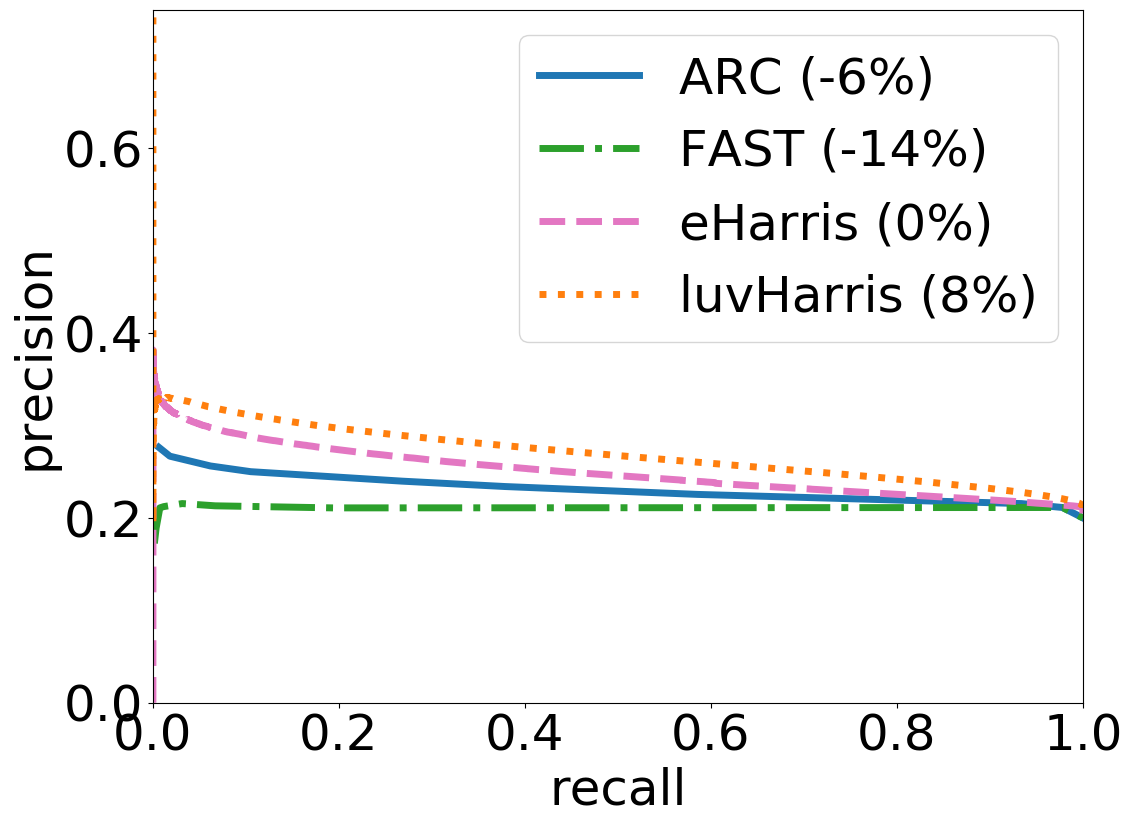}
	\includegraphics[width=0.23\linewidth]{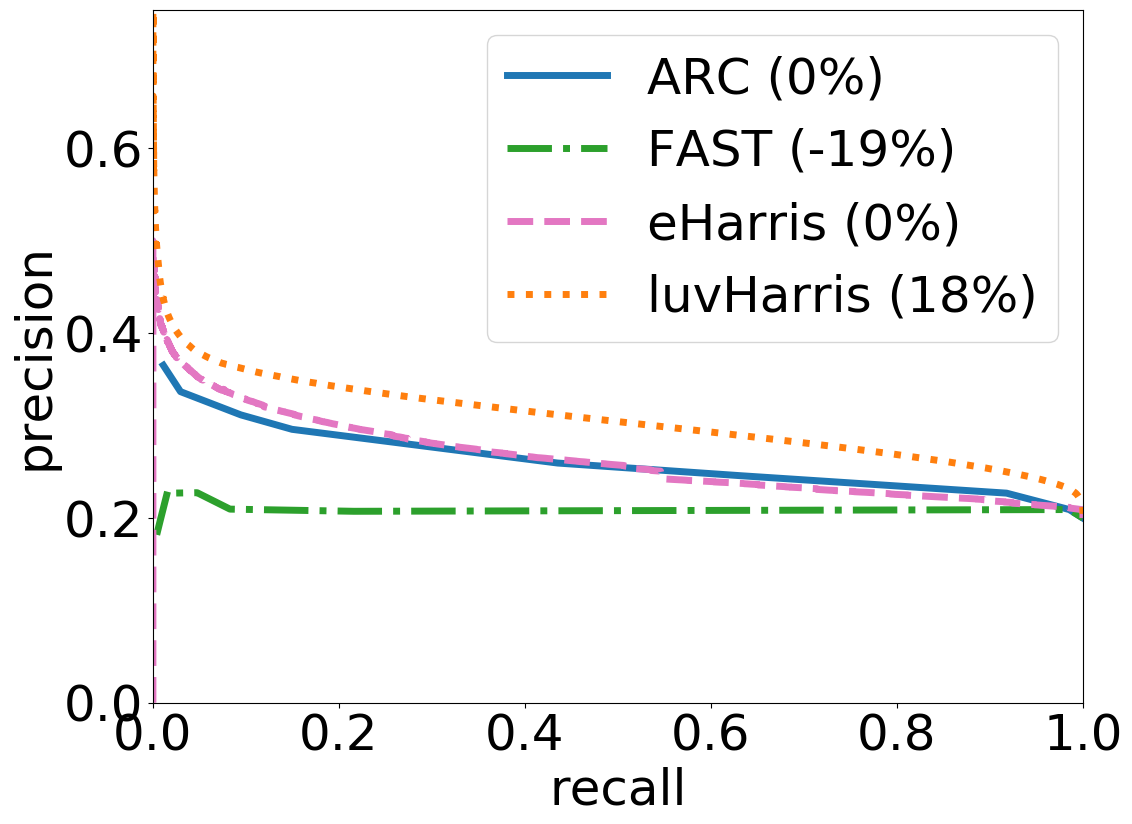}
	\includegraphics[width=0.23\linewidth]{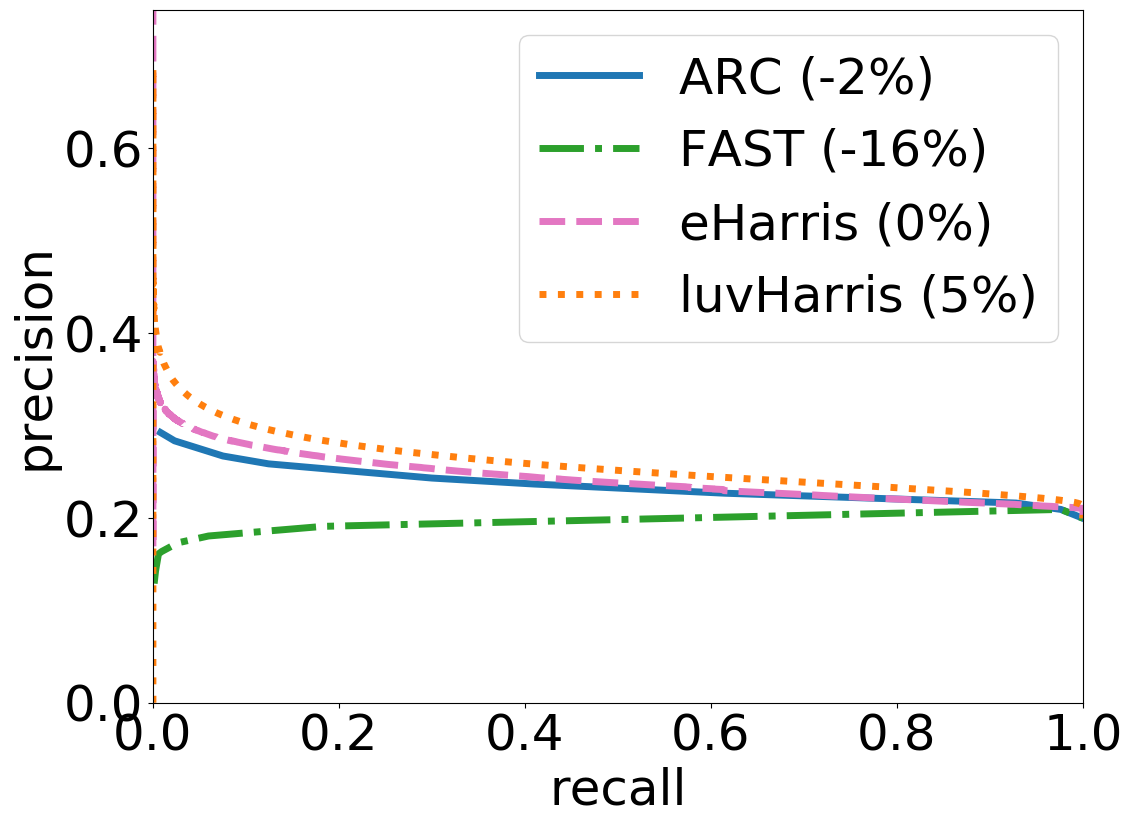}
	\includegraphics[width=0.23\linewidth]{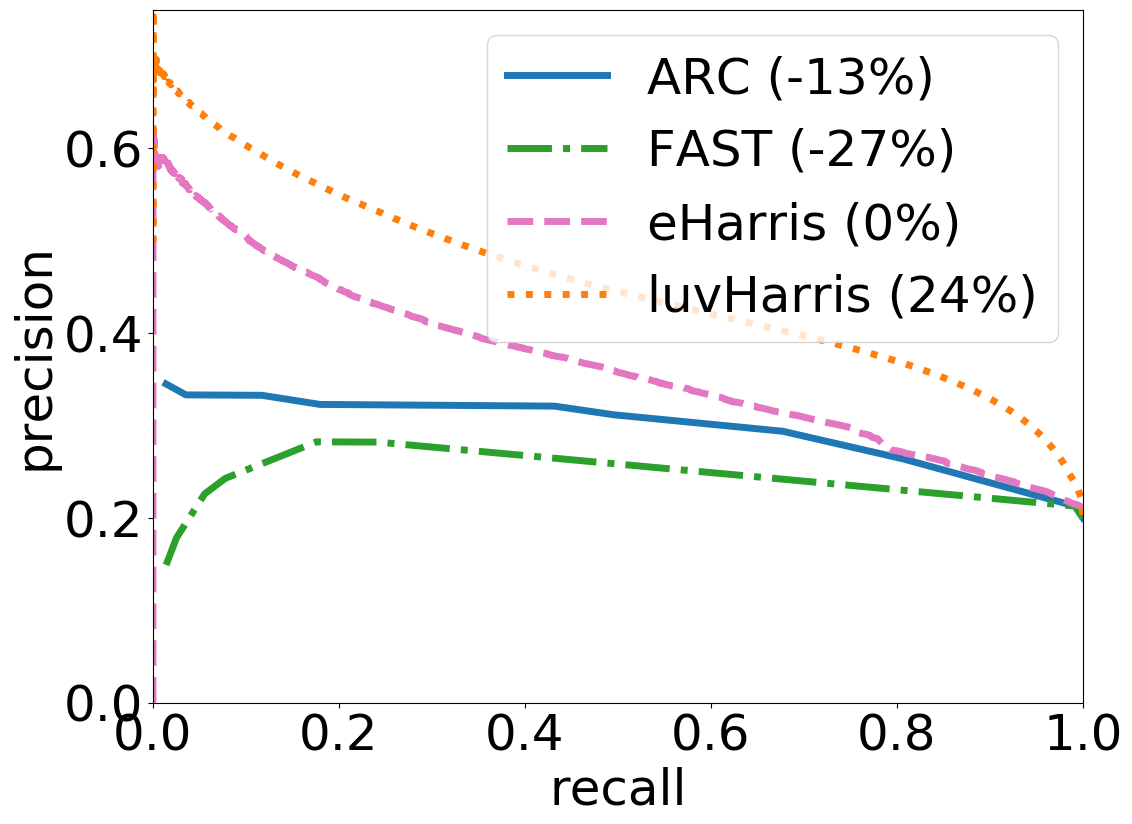}}\\

	\caption{Examples and detection accuracy results for the (left to right) \texttt{boxes\_6dof, dynamic\_6dof, poster\_6dof}, and \texttt{shapes\_6dof} datasets~\cite{EDdatasets}. \hl{Legend boxes indicate the percentage improvement over the eHarris baseline for 50\% recall. A perfect detection algorithm has a 1.0 precision and a 1.0 recall, and therefore good algorithms should push towards the top right corner of the results plots.}}
	\label{fig:main_result}
\end{figure*}

\paragraph*{Ground-truth}corners\footnote{used as the basis for analysis, we acknowledge the method is not infallible. The task of deciding which events are corners will always be somewhat ambiguous.} were computed for all datasets by first creating a high temporal resolution intensity image video using the e2vid~\cite{e2vid} algorithm. The ground-truth corner score was then assigned to each event using the output of the OpenCV \texttt{cornerHarris} method applied to the temporally closest frame. The Harris algorithm is a widely accepted baseline in the field of computer vision. Finally, a threshold was applied to the scores to result in the set of true corner events, with an assumption that \hl{the 20\% highest scoring of all events in a dataset were corner events}. Intensity frames were generated every 1000-3000 events depending on the texture in the dataset. \hl{For example, \texttt{boxes\_6dof} has more texture than \texttt{shapes\_6dof} and requires a larger batch size for e2vid, but the percentage of corners in a scene is assumed to remain constant between datasets for all experiments.} From a qualitative assessment, \hl{any error from processing the ground-truth in batches} was insignificant compared to other possible sources of error, such as artefacts in the frame reconstruction. Fig.~\ref{fig:main_gt} shows an example e2vid frame for each dataset, as well as the slice of events associated with the single frame, indicating those classified as corners by the original Harris algorithm running on the reconstructed frames.

\hl{The assumption of 20\% corners is not perfect as shapes with different angle compositions will have different percentage of corners. However neither could a fixed ground-truth threshold be used across all datasets as the average image intensity was not consistent across datasets, i.e. the output of e2vid was not always consistent. The fixed percentage was therefore chosen rather than arbitrarily setting individual ground-truth thresholds for each dataset. The datasets were afterwards verified qualitatively, to ensure the generated ground-truth was valid, as can be confidently seen in} Fig.~\ref{fig:main_gt}.

\paragraph*{Precision-recall} plots show algorithm accuracy as each algorithm's corner decision parameter is varied. For example, an algorithm with a high threshold will result in only a few corners (low recall), but should have less false positive\footnote{\hl{a false positive occurs when the algorithm classifies the event as a corner but is not a corner according to the ground-truth}} detections (high precision). The precision-recall metric allows a better comparison between algorithms which use different parameters for the corner decision. FAST and ARC algorithms use the \textit{arc length} for the decision parameter; as the inclusion angle is swept from strictly $90^{\circ}$ to anything in a $180^{\circ}$ arc the algorithm goes from high-precision to high-recall. Instead, eHarris and luvHarris use the corner score from the Harris equation. 

Fig.~\ref{fig:main_pr} shows that in almost all cases the FAST algorithm results in the lowest performance as it, compared to eHarris, trades off accuracy for faster computation~\cite{Mueggler2017a}. FAST also has no mechanism for $270^{\circ}$ corners such that it misses possible detections. In many datasets, the ARC algorithm achieves an accuracy on-par with eHarris. As both are able to detect $270^{\circ}$ corners it further supports the hypothesis of such a problem with FAST. 

The comparison of non-filtered (Fig.~\ref{fig:main_prf}) and filtered ARC (Fig.~\ref{fig:main_pr}), \hl{especially for the \texttt{shapes\_6dof} dataset, shows the necessity of the filter for the algorithm. Additionally the Harris-based algorithms also experience an accuracy improvement, using the filtered dataset. The spatially-adaptive region of eHarris and the TOS of luvHarris become cleaner with the filtered event input giving rise to more consistent gradients. However, the filter is not \textit{necessary} for the algorithms, which allows for more flexibilty of use. The superior accuracy of the Harris-based algorithms is clear on the \texttt{shapes\_6dof} dataset, given the reasons explained in} Section~\ref{sec:arcissues}.

The luvHarris algorithm improves the accuracy of corner detection over eHarris, substantially in most datasets.  

\paragraph*{Further analysis} the algorithm precision was compared at the 50\% recall mark. That is, the decision threshold was tuned to detect 50\% of the ground-truth corner events for all algorithms, and the number of correct corners were compared. The \hl{legend} boxes in Fig.~\ref{fig:main_pr} and Fig.~\ref{fig:main_prf} indicate the percentage improvement over the eHarris algorithm. In all cases, luvHarris shows the biggest precision improvement, up to a 40\%.

\begin{figure}
    \centering
	\subcaptionbox{\texttt{shapes\_6dof}}{
	\includegraphics[width=0.7\linewidth]{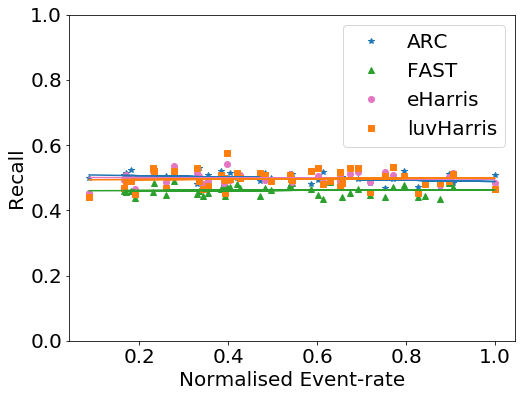}}\\
	\subcaptionbox{\texttt{dynamic\_6dof}}{
	\includegraphics[width=0.7\linewidth]{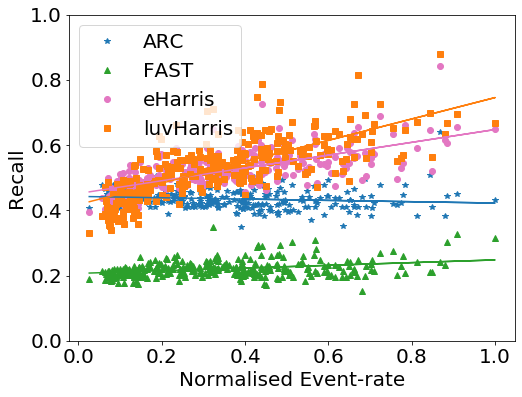}}
	\caption{\hl{Recall rates compared to event-rate for the (a) \texttt{shapes\_6dof} and (b) \texttt{dynamic\_6dof} datasets. A threshold that produced closest to 50\% recall was selected for each algorithm (but not possible to achieve exactly). The measured update rate of luvHarris Algorithm}~\ref{alg:harris} \hl{was approximately 1 kHz.} while other algorithms are event-by-event.}
	\label{fig:recallvsrate}
\end{figure}

\hl{At the 50\% recall mark, all algorithms produce a corner recall that is constant compared to the event-rate of the dataset. Such a result is expected for event-by-event algorithms as each event is processed identically, but can affect luvHarris as updates of $L$ are not performed for each and every event. The update rate of }Algorithm~\ref{alg:harris} \hl{was approximately 1 kHz and was sufficient for consistent recall on the \texttt{shapes\_6dof} dataset, as shown in} Fig.~\ref{fig:recallvsrate}. For the \texttt{dynamic\_6dof} \hl{The eHarris and luvHarris algorithms showed slight upwards trends for recall at faster event-rates which would require further investigation to understand the particular cause.}

\begin{figure}
    \centering
	\subcaptionbox{Non-filtered events\label{fig:ablation_nf}}{
	\includegraphics[width=0.7\linewidth]{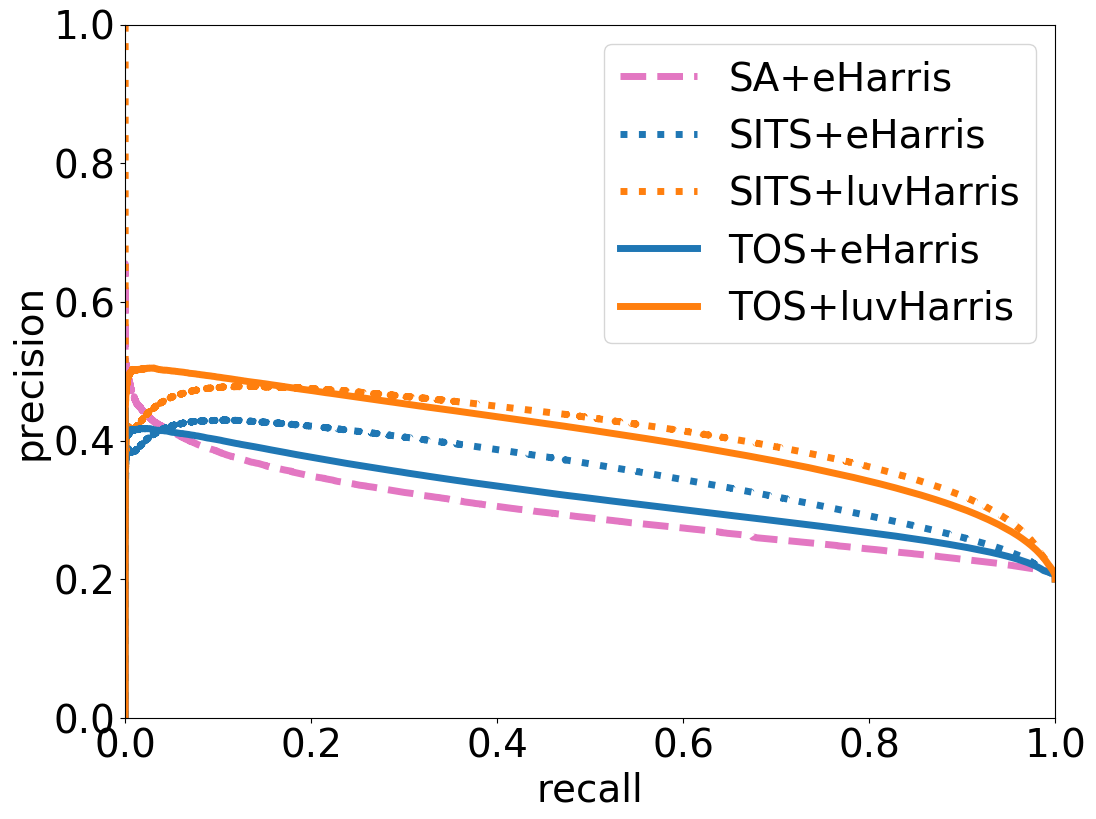}}\\
	\subcaptionbox{Filtered Events\label{fig:ablation_f}}{
	\includegraphics[width=0.7\linewidth]{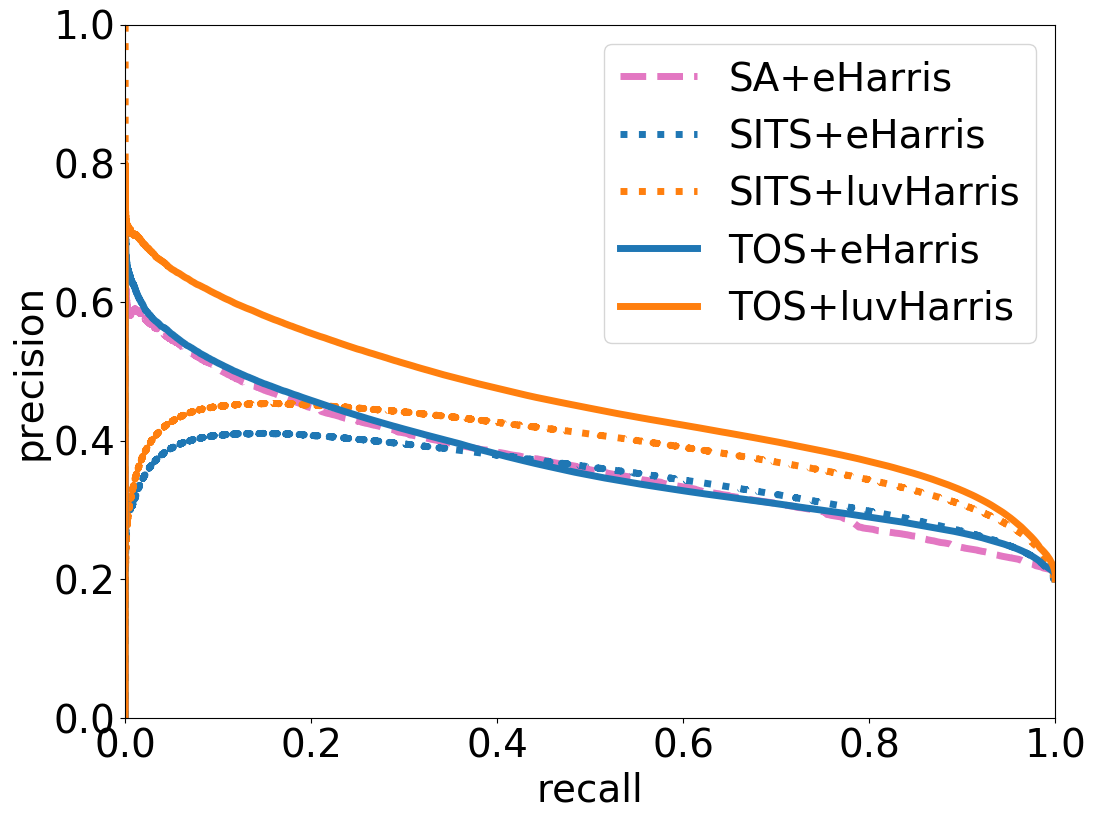}}
	\caption{\hl{A comparison of Harris method (eHarris, luvHarris) and surface methods (SITS, TOS, Spatially-Adaptive) for corner accuracy on the \texttt{shapes\_6dof} dataset. The Spatially-Adaptive surface is incompatible with luvHarris.}}
	\label{fig:ablation}
\end{figure}

\hl{Computationally the SITS and TOS surface require almost identical computations, and are therefore exchangeable in a real-time system.} Fig.~\ref{fig:ablation} \hl{shows a comparison of luvHarris and eHarris, in terms of accuracy, when using different surface types. , the results indicate:}
\begin{itemize}
    \item \hl{The SITS surface is tolerant to the raw event-stream as the event-filter only marginally improved the performance, i.e. comparing SITS curves between} Fig.~\ref{fig:ablation_nf} and Fig.~\ref{fig:ablation_f}.
    \item \hl{The TOS filter performs significantly better in conjunction with the event-filter.}
    \item \hl{The TOS surface with the event-filter behaves very similarly to the spatially-adaptive surface as seen by the overlapping eHarris curves }in Fig.~\ref{fig:ablation_f}, i.e. the TOS keeps the most recent $k_{TOS}$ pixels in any image patch. The TOS surface is cheaper to compute than the spatially-adaptive method.
    \item \hl{The best performance is achieved with the TOS and luvHarris with the event-filter.}
    \item \hl{luvHarris performs significantly better than eHarris independently of the underlying surface used.}
\end{itemize}

\subsection{Qualitative corner quality}

\begin{figure}
	\centering
	\subcaptionbox{ARC}{
	\includegraphics[width=0.45\linewidth]{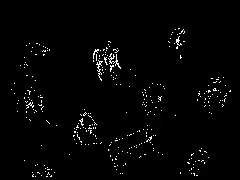}}
	\subcaptionbox{eHarris}{
	\includegraphics[width=0.45\linewidth]{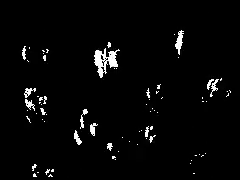}}\\
	\subcaptionbox{luvHarris}{
	\includegraphics[width=0.45\linewidth]{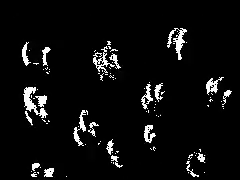}}
	\subcaptionbox{FAST}{
	\includegraphics[width=0.45\linewidth]{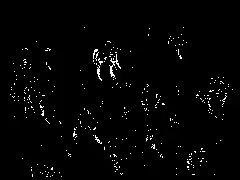}}
	\caption{\hl{Synchronised} qualitative corner trails over a 100 ms window for the \texttt{shapes\_6dof} dataset. (b, c) Harris based algorithms produce consistent, wider trails, while (a, d) arc-based algorithms are more affected by missed corner events, and falsely classifying edges as corners.}
	\label{fig:shapes_qual}
\end{figure}

\begin{figure}
	\centering
	\subcaptionbox{ARC}{
	\includegraphics[width=0.45\linewidth]{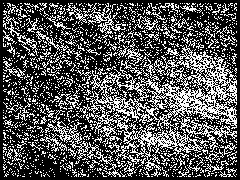}}
	\subcaptionbox{eHarris}{
	\includegraphics[width=0.45\linewidth]{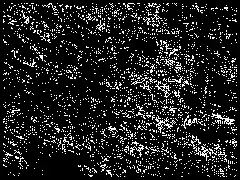}}\\
	\subcaptionbox{luvHarris}{
	\includegraphics[width=0.45\linewidth]{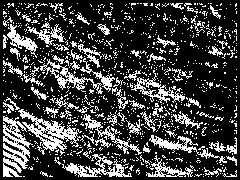}}
	\subcaptionbox{FAST}{
	\includegraphics[width=0.45\linewidth]{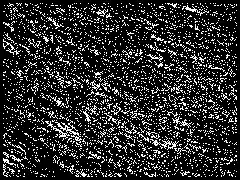}}
	\caption{\hl{Synchronised} qualitative corner trails over a 100 ms window for the \texttt{boxes\_6dof} dataset. In cluttered conditions exactly what is a ``corner'' is more ambiguous, however it is clear that (c) luvHarris detects somewhat consistent trails, while (a) ARC and (d) FAST do not.}
	\label{fig:boxes_qual}
\end{figure}

\begin{figure}
	\centering
	\subcaptionbox{[@12s, 3M events/s]\label{fig:qual1}}
	{\includegraphics[width=0.9\linewidth]{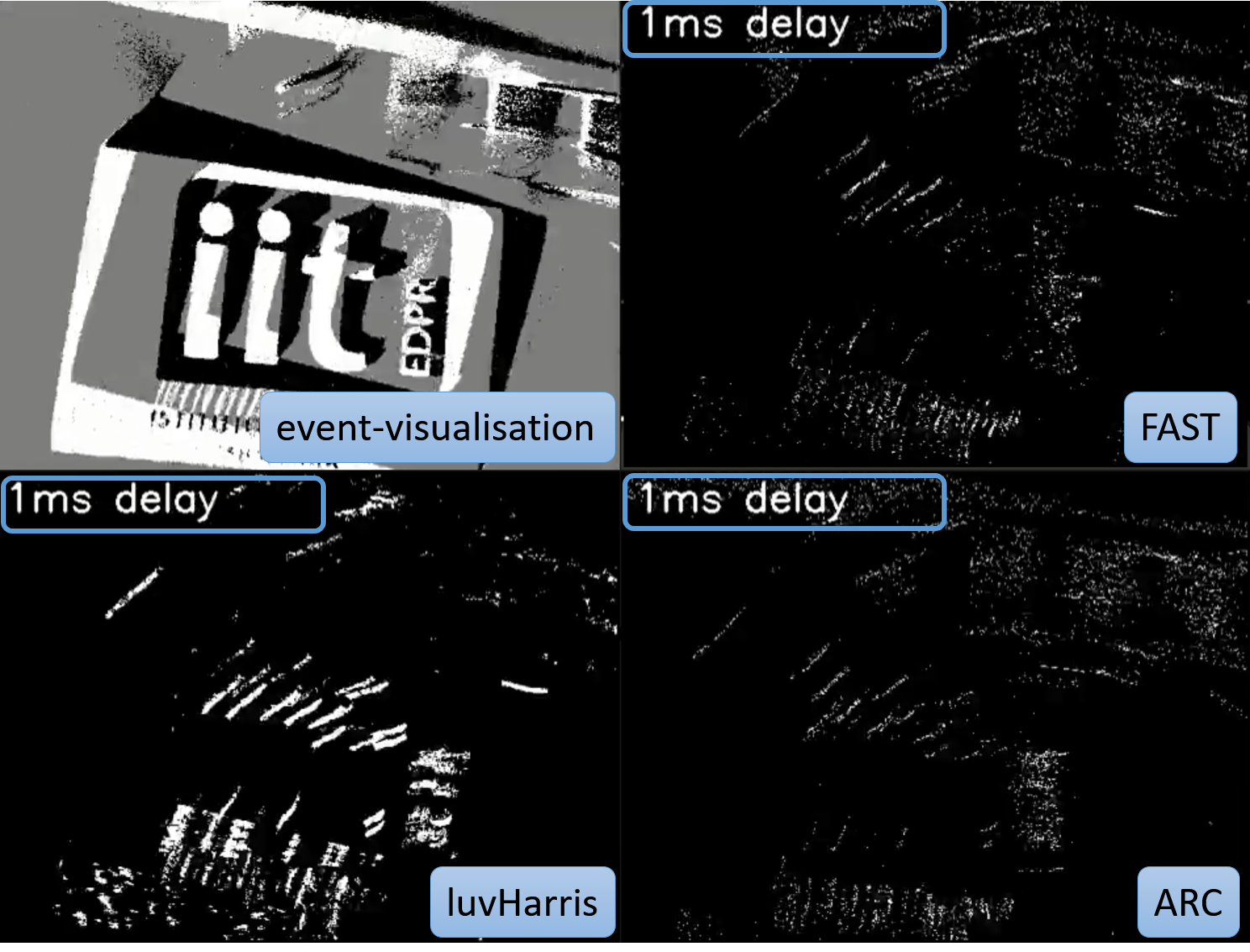}}	
	\subcaptionbox{[@21s, 5M events/s]\label{fig:qual2}}
	{\includegraphics[width=0.9\linewidth]{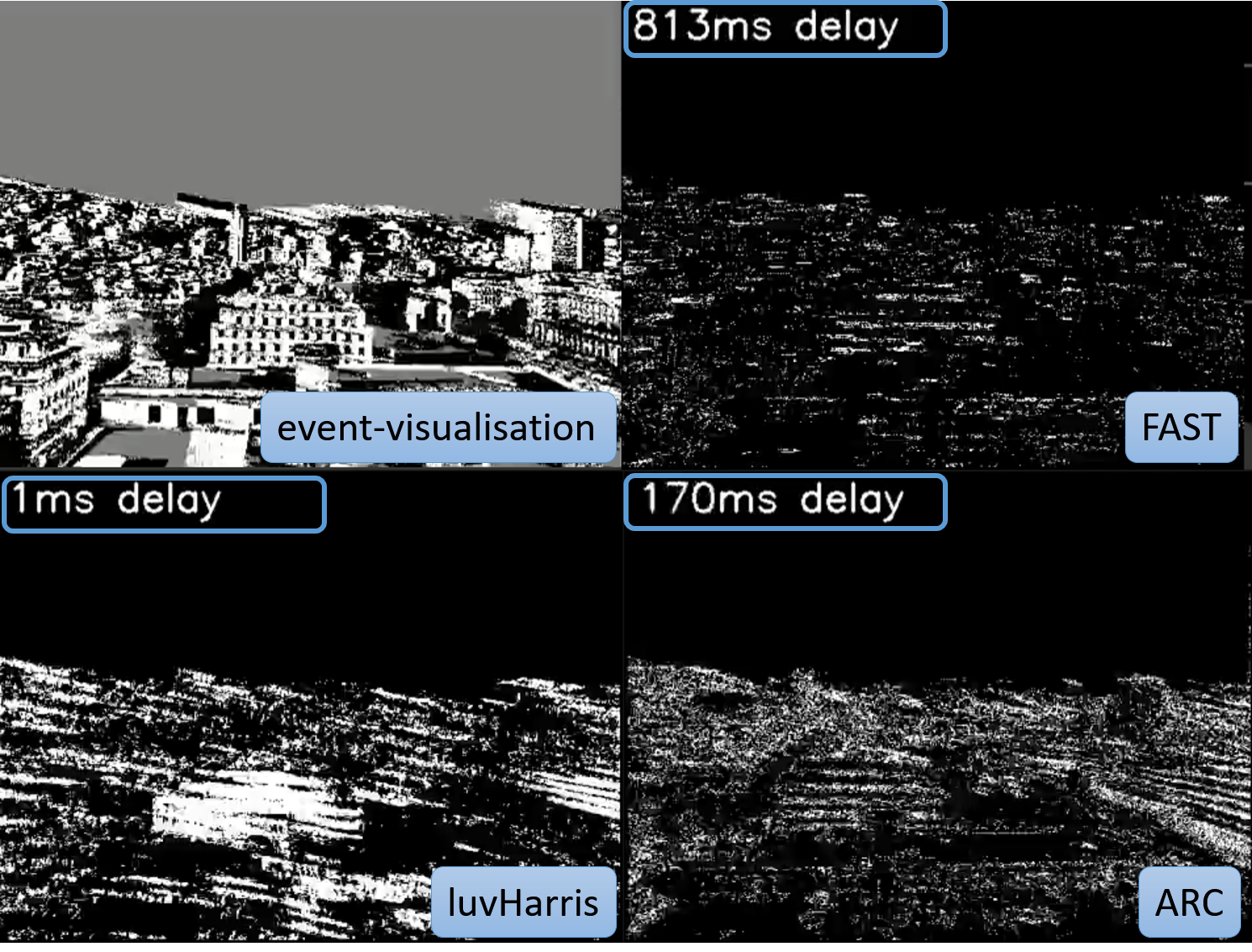}}	
	\subcaptionbox{[@28s, 8M events/s]\label{fig:qual4}}
	{\includegraphics[width=0.9\linewidth]{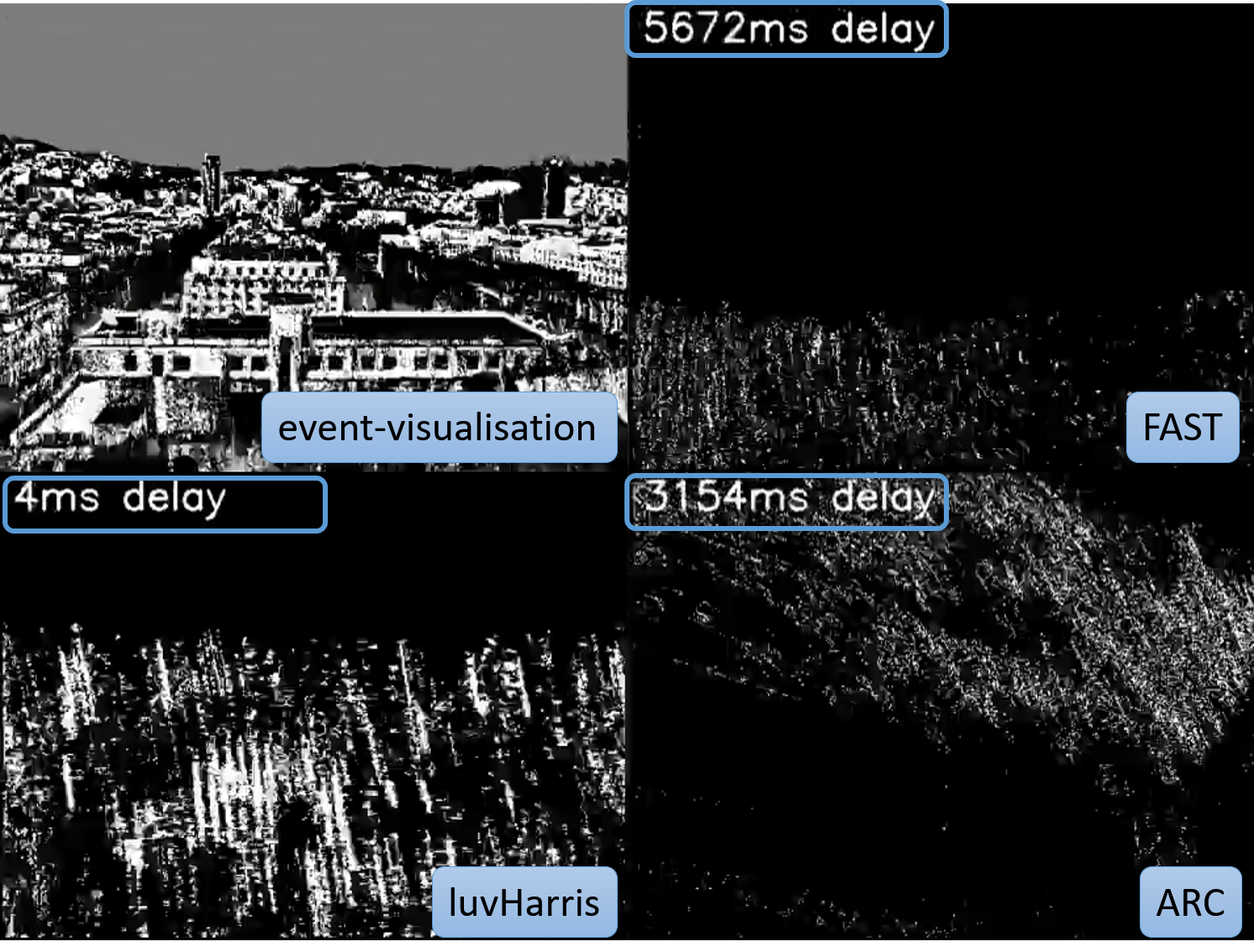}}
	\caption{An \hl{un-synchronised} qualitative visualisation over a 100 ms compute time of luvHarris, ARC, and FAST run on-line simultaneously and delay accumulated by the algorithm is stated. All algorithms are selective to corners in simple scenes, but in complex scenes ARC and FAST produce more false positives and less consistent detections over time compared with luvHarris. eHarris was not run for the on-line experiment as it was too computationally heavy. The result is best seen in \href{https://zenodo.org/record/4739290}{video} format.}
	\label{fig:qualitative}
\end{figure}

Fig.~\ref{fig:shapes_qual} and ~\ref{fig:boxes_qual} show corner traces for the \texttt{shapes\_6dof} and \texttt{boxes\_6dof} datasets respectively, at a \hl{fixed synchronised point in time}. Fig.~\ref{fig:qualitative} \hl{shows multiple stills of the live experiment, which is specifically not synchronised to indicate not only the corner quality, but the delays in the algorithm when operating live, under high event-rate conditions.}

In simple scenes, e.g. Fig.~\ref{fig:shapes_qual} and Fig.~\ref{fig:qual1}, all algorithms are selective to corners, which validates a correct implementation of the algorithms. However, in cluttered scenes the selectivity of all algorithms to corners is questionable. Fig.~\ref{fig:qual2} shows that FAST misses many corners, and Fig.~\ref{fig:qual4} shows that, as expected, ARC produces many false positives, given it is designed to favour false positives, over missed detections~\cite{Alzugaray2018}, such that finally tracking can filter out the false positive detections. However, in Fig.~\ref{fig:boxes_qual} both ARC and FAST produce noisy responses such that it is hard to discern any consistent corner traces that could be tracked well. Instead, luvHarris provides consistent selective detections over time, in addition to some noisy detections. 

\section{DISCUSSION} \label{section:discuss}

\paragraph*{Useability} from practical assessment of the previous state-of-the-art with live camera data, it was clear that the arc-based methods simply did not produce strong and consistent corner detections, which is shown in the qualitative results. Notably, ARC produced too many false positives for useful results, and it is arguable whether a further corner tracking layer would function well. Indeed, ARC was designed in conjunction with a corner tracker, but we argue that better underlying detections will always also produce better tracking. Instead, the low event-throughput of eHarris made it unsuitable for higher-resolution cameras. Quantitatively, luvHarris improved over the state-of-the-art in both accuracy and event-throughput, but more importantly, the qualitative output shows consistent trails of corners that should be more easily tracked for future motion estimation tasks. 

Achieving real-time for the HVGA resolution is an improvement over previous state-of-the-art, however the maximum event-rate of luvHarris might still be less than satisfactory for even higher resolution cameras. Considering \hl{our} results, it may actually be impossible to perform event-by-event computation for such cameras without specific neuromorphic hardware.

\paragraph*{Comparison to literature} a recent comparison of event-based corner detectors~\cite{yilmaz2021evaluation} similarly concludes that ARC was the only real-time detector on the RPG datasets, but did not test higher-resolution event-cameras such as the ATIS generation 3. Unfortunately, they only present true positive rate as an accuracy metric, which does not give a full understanding of the performance as the precision-recall curves. Instead, \cite{Alzugaray2018} presented both true and false positive rates, and similarly indicate that more than 50\% precision is not expected for event-based algorithms on RPG datasets. Previous literature stating true and false positive rates are selecting a single point along the precision-recall curve to display as a result. We instead propose that the full curve gives a better overall picture of the true algorithm accuracy. The exact value is, however, highly-dependent on thresholds used to classify ground-truth. 

\hl{The result that luvHarris outperformed eHarris on accuracy, when using identical surface methods was surprising. It could be assumed that the two outputs would be identical. Our hypothesis is that testing for corners at an event-level temporal resolution could introduce noise in the corner spatial pattern, which is smoothed by processing at the lower rate of Algorithm}~\ref{alg:harris}. \hl{Alternatively, convolutional edge cases when computing over the image patch may affect the eHarris computation, which does not occur when computing over then entire image as in luvHarris.}

\paragraph*{Dataset validity} from our results, the RPG \texttt{boxes\_6dof} and \texttt{poster\_6dof} are questionable datasets to use for quantitative comparison of corner detection. The flat precision-recall curves obtained indicate the algorithms are not performing much better than chance - however it can be seen that on \texttt{shapes\_6dof} the algorithms \textit{are indeed} selective to corners. Such a result could indicate an incorrect ground-truth, but we suggest the datasets are too cluttered to be easily used to measure performance. For example, it does not matter exactly what an algorithm decides is a `corner' as long as it consistently selects the same position over time. In such cluttered datasets, it is hard to determine precisely a ground-truth of corner and not-corner. \hl{Metrics that require a tracking layer as presented in}~\cite{yilmaz2021evaluation} \hl{may be better for these datasets.} Indeed, the results should also be slightly biased towards the luvHarris on these datasets as the ground-truth also uses the \texttt{cv::cornerHarris} method. The same `type' of corners may be selected despite the very different images used for ground-truth and luvHarris. While we suggest the quantitative results are questionable, identifying consistent detections over-time in these datasets is still important, as we show in Fig.~\ref{fig:boxes_qual}.

\paragraph*{Known issues} \hl{event-by-event algorithms are more computationally efficient (use less resources for the same result) than luvHarris for low-rate event-streams. In such situations, the computation over the full retina results in redundant computation, as opposed to only processing change events. However, it could also be possible to further improve luvHarris and dynamically throttle} Algorithm~\ref{alg:harris} during periods of little motion.


The TOS is still defined by a parameter, $T_{TOS}$.  However, $T_{TOS}$ is defined by the \textit{application} rather than external conditions (e.g. object speed). In this case, we set $T_{TOS}$ to give a clear edge that promotes corner detection. \hl{The SITS instead of the TOS could be a reasonable choice as its performance didn't degrade using the raw event-stream (no event-filter). Removing the event-filter removes several parameters that must be tuned to the application and conditions. For maximum accuracy, the TOS and event-filter should be used.}

\paragraph*{Event-by-event v.s. batch} an ongoing discussion occurs around the validity of batch computation for event-based cameras. The evidence in this study suggests that, for CPU processing of high-resolution cameras, only very limited processing can be performed for \textit{every single event.} Indeed, the bottle-neck of luvHarris was still the event-by-event TOS update per event, rather than the rate of \texttt{cv::cornerHarris} over the full retina. We therefore propose that the hybrid concept presented in this paper offers a good compromise between the two - complex algorithmic computations are performed in batch, while the event-stream is still read and output asynchronously. The events flow in and out just as in any fully event-by-event algorithm. On dedicated neuromorphic hardware fully event-by-event algorithms can still be realised.

\section{CONCLUSION} \label{section:conc}

We have presented a practical corner detector for event-based camera specifically addressing problems with limits on event-throughput and detector accuracy. Accuracy is improved by using the Harris algorithm; compared to the arc-based methods it uses more information to give a consistent result. The consistent corner trails even in cluttered conditions indicate the luvHarris algorithm will also produce consistent motion estimation when tracking corners over time. 

Compared to previous event-driven Harris implementations, we use the proposed threshold-ordinal-surface which eliminates the need for a temporal parameter, and has a simple update methodology. The contribution of the TOS also extends beyond the corner detection algorithm, and the surface could also be used in other applications. The full luvHarris algorithm is simple and can be implemented in very few lines of code, and builds on open-source, optimised, libraries.

Event-throughput for high-resolution cameras of multiple million events/s was achieved in real-time by decoupling the heavy Harris calculations from the event-stream. Instead, the only calculations that were done event-by-event was the update of the TOS and a simple look-up of the best effort Harris score. The concept of decoupling event streams from the complex algorithm component is a valuable insight that sits somewhere between fully asynchronous and batch-based processing and can be applied to other event-driven vision algorithms; to enable real-time event-based algorithms, possibly an approach using look-up table events (luv) is all you need.

\addtolength{\textheight}{-1.6cm}
\bibliography{bibliography}

\begin{thebibliography}{10}
\providecommand{\url}[1]{#1}
\csname url@samestyle\endcsname
\providecommand{\newblock}{\relax}
\providecommand{\bibinfo}[2]{#2}
\providecommand{\BIBentrySTDinterwordspacing}{\spaceskip=0pt\relax}
\providecommand{\BIBentryALTinterwordstretchfactor}{4}
\providecommand{\BIBentryALTinterwordspacing}{\spaceskip=\fontdimen2\font plus
\BIBentryALTinterwordstretchfactor\fontdimen3\font minus
  \fontdimen4\font\relax}
\providecommand{\BIBforeignlanguage}[2]{{%
\expandafter\ifx\csname l@#1\endcsname\relax
\typeout{** WARNING: IEEEtran.bst: No hyphenation pattern has been}%
\typeout{** loaded for the language `#1'. Using the pattern for}%
\typeout{** the default language instead.}%
\else
\language=\csname l@#1\endcsname
\fi
#2}}
\providecommand{\BIBdecl}{\relax}
\BIBdecl

\bibitem{luo2013survey}
Z.~Luo, ``Survey of corner detection techniques in image processing,''
  \emph{International Journal of Recent Technology and Engineering (IJRTE)},
  vol.~2, no.~2, pp. 184--185, 2013.

\bibitem{Vasco2017}
V.~Vasco, A.~Glover, E.~Mueggler, D.~Scaramuzza, L.~Natale, and C.~Bartolozzi,
  ``{Independent motion detection with event-driven cameras},'' \emph{2017 18th
  International Conference on Advanced Robotics, ICAR 2017}, no. July, pp.
  530--536, 2017.

\bibitem{Vidal2018}
\BIBentryALTinterwordspacing
A.~R. Vidal, H.~Rebecq, T.~Horstschaefer, and D.~Scaramuzza, ``{Ultimate SLAM?
  Combining Events, Images, and IMU for Robust Visual SLAM in HDR and High
  Speed Scenarios},'' \emph{IEEE Robotics and Automation Letters}, 2018.
  [Online]. Available:
  \url{http://arxiv.org/abs/1709.06310{\%}0Ahttp://dx.doi.org/10.1109/LRA.2018.2793357}
\BIBentrySTDinterwordspacing

\bibitem{Vasco2016a}
\BIBentryALTinterwordspacing
V.~Vasco, A.~Glover, and C.~Bartolozzi, ``{Fast event-based Harris corner
  detection exploiting the advantages of event-driven cameras},'' \emph{2016
  IEEE/RSJ International Conference on Intelligent Robots and Systems (IROS)},
  pp. 4144--4149, 2016. [Online]. Available:
  \url{http://ieeexplore.ieee.org/document/7759610/}
\BIBentrySTDinterwordspacing

\bibitem{Mueggler2017a}
E.~Mueggler, C.~Bartolozzi, and D.~Scaramuzza, ``{Fast Event-based Corner
  Detection},'' \emph{British Machine Vision Conference}, vol.~1, pp. 1--11,
  2017.

\bibitem{Alzugaray2018}
\BIBentryALTinterwordspacing
I.~Alzugaray and M.~Chli, ``{Asynchronous Corner Detection and Tracking for
  Event Cameras in Real-Time},'' \emph{IEEE Robotics and Automation Letters},
  vol.~3, no.~4, pp. 3177--3184, 2018. [Online]. Available:
  \url{https://ieeexplore.ieee.org/document/8392795/}
\BIBentrySTDinterwordspacing

\bibitem{Li2019}
R.~{Li}, D.~{Shi}, Y.~{Zhang}, K.~{Li}, and R.~{Li}, ``Fa-harris: A fast and
  asynchronous corner detector for event cameras,'' in \emph{2019 IEEE/RSJ
  International Conference on Intelligent Robots and Systems (IROS)}, 2019, pp.
  6223--6229.

\bibitem{Manderscheid2019}
J.~{Manderscheid}, A.~{Sironi}, N.~{Bourdis}, D.~{Migliore}, and V.~{Lepetit},
  ``Speed invariant time surface for learning to detect corner points with
  event-based cameras,'' in \emph{2019 IEEE/CVF Conference on Computer Vision
  and Pattern Recognition (CVPR)}, 2019, pp. 10\,237--10\,246.

\bibitem{harris1988combined}
C.~G. Harris, M.~Stephens \emph{et~al.}, ``A combined corner and edge
  detector.'' in \emph{Alvey vision conference}, vol.~15, no.~50.\hskip 1em
  plus 0.5em minus 0.4em\relax Citeseer, 1988, pp. 10--5244.

\bibitem{opencv_library}
G.~Bradski, ``{The OpenCV Library},'' \emph{Dr. Dobb's Journal of Software
  Tools}, 2000.

\bibitem{Lichtsteiner2008}
\BIBentryALTinterwordspacing
P.~Lichtsteiner, C.~Posch, and T.~Delbruck, ``{A 128x128 120 dB 15 us latency
  Asynchronous Temporal Contrast Vision Sensor},'' \emph{IEEE Journal of
  Solid-State Circuits}, vol.~43, no.~2, pp. 566--576, 2008. [Online].
  Available:
  \url{http://ieeexplore.ieee.org/xpls/abs{\_}all.jsp?arnumber=4444573}
\BIBentrySTDinterwordspacing

\bibitem{Posch2008}
C.~Posch, D.~Matolin, and R.~Wohlgenannt, ``{An asynchronous time-based image
  sensor},'' in \emph{2008 IEEE International Symposium on Circuits and
  Systems}, may 2008, pp. 2130--2133.

\bibitem{Brandli2014}
C.~{Brandli}, R.~{Berner}, M.~{Yang}, S.~{Liu}, and T.~{Delbruck}, ``A 240 x
  180 130 db 3 us latency global shutter spatiotemporal vision sensor,''
  \emph{IEEE Journal of Solid-State Circuits}, vol.~49, no.~10, pp. 2333--2341,
  Oct 2014.

\bibitem{Scheerlinck}
C.~{Scheerlinck}, N.~{Barnes}, and R.~{Mahony}, ``Asynchronous spatial image
  convolutions for event cameras,'' \emph{IEEE Robotics and Automation
  Letters}, vol.~4, no.~2, pp. 816--822, 2019.

\bibitem{chiberre2021detecting}
P.~Chiberre, E.~Perot, A.~Sironi, and V.~Lepetit, ``Detecting stable keypoints
  from events through image gradient prediction,'' in \emph{Proceedings of the
  IEEE/CVF Conference on Computer Vision and Pattern Recognition}, 2021, pp.
  1387--1394.

\bibitem{EDdatasets}
E.~Mueggler, H.~Rebecq, G.~Gallego, T.~Delbruck, and D.~Scaramuzza, ``The
  event-camera dataset and simulator: Event-based data for pose estimation,
  visual odometry, and slam,'' \emph{The International Journal of Robotics
  Research}, vol.~36, no.~2, pp. 142--149, 2017.

\bibitem{yilmaz2021evaluation}
{\"O}.~Y{\i}lmaz, C.~Simon-Chane, and A.~Histace, ``Evaluation of event-based
  corner detectors,'' \emph{Journal of Imaging}, vol.~7, no.~2, p.~25, 2021.

\bibitem{e2vid}
H.~{Rebecq}, R.~{Ranftl}, V.~{Koltun}, and D.~{Scaramuzza}, ``Events-to-video:
  Bringing modern computer vision to event cameras,'' in \emph{2019 IEEE/CVF
  Conference on Computer Vision and Pattern Recognition (CVPR)}, 2019, pp.
  3852--3861.

\end{thebibliography}

\begin{IEEEbiography}[{\includegraphics[width=1in,height=1.25in,clip,keepaspectratio]{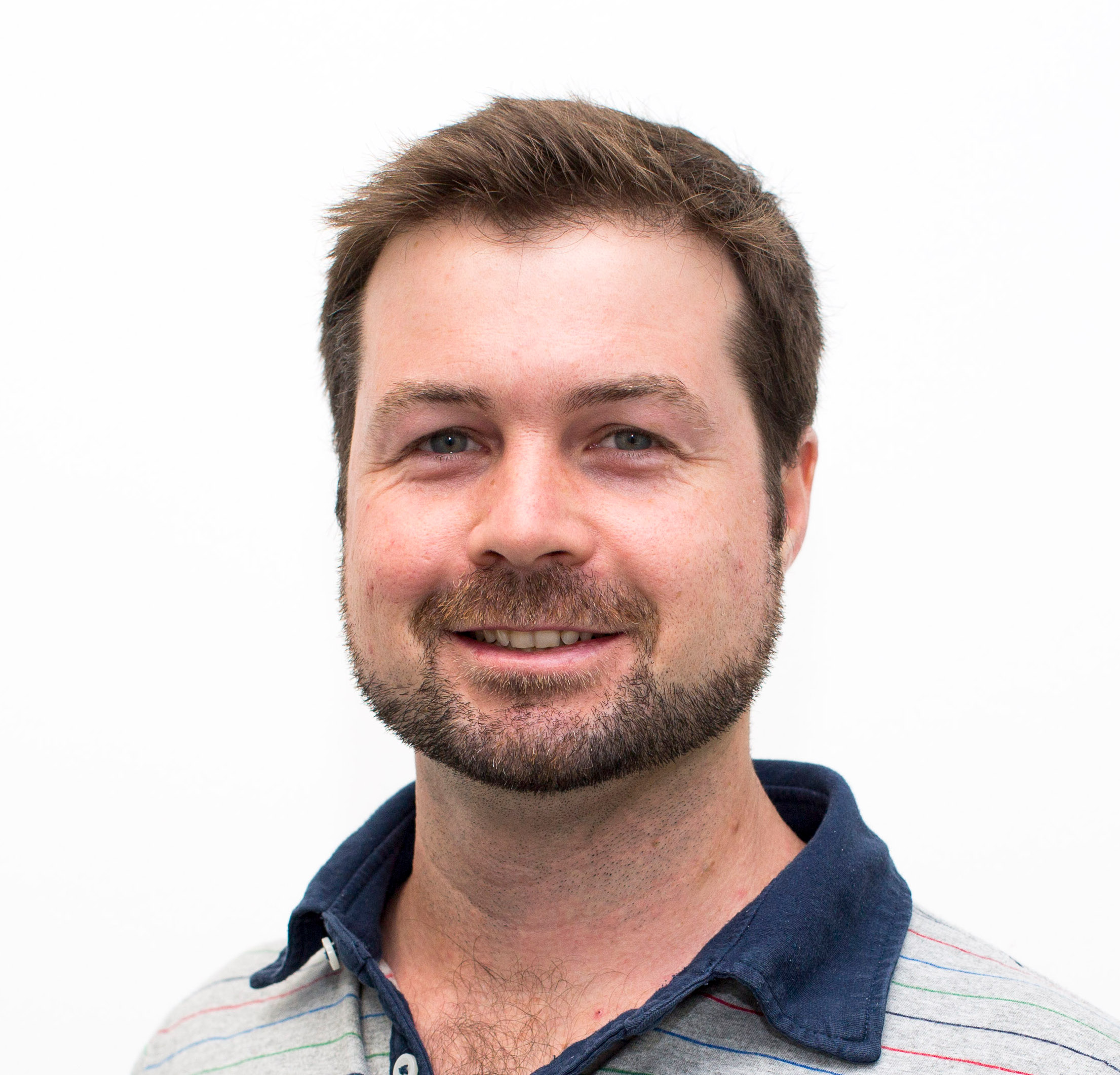}}]{Arren Glover}
Arren obtained a Bachelor of Mechatronic Engineering from the University of Queensland and a PhD in developmental robotics and on-line learning from the Queensland University of Technology, Australia. Since 2015 he has been with the Italian Institute of Technology's Event-driven Perception for Robotics research line with the focus on developing real-time vision algorithms for the iCub robot using event-cameras. He is a senior researcher also leading industrial collaborations.
\end{IEEEbiography}
\begin{IEEEbiography}[{\includegraphics[width=1in,height=1.25in,clip,keepaspectratio]{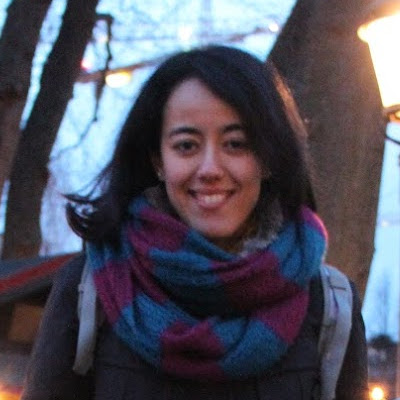}}]{Aiko Dinale}
Aiko Dinale received the B.E. degree in 2010 in computer engineering from the University of Genova (Italy). In 2012, she received the M.E. degree in robotics engineering jointly from Ecole Centrale de Nantes (France) and University of Genoa (Italy) as Erasmus Mundus scholarship holder for the European Master on Advanced Robotics (EMARO). In 2017, she earned her PhD title in Mechanical Engineering (Robotics and Mechatronics curricula) from the University of Genoa (Italy). Since then, she is a postdoc at the Italian Institute of Technology (IIT) in Genoa (Italy). She has been working on various topics such as whole-body dynamics, and control of humanoid robots, and event-driven perception systems for robotics.
\end{IEEEbiography}


\begin{IEEEbiography}[{\includegraphics[width=1in,height=1.25in,clip,keepaspectratio]{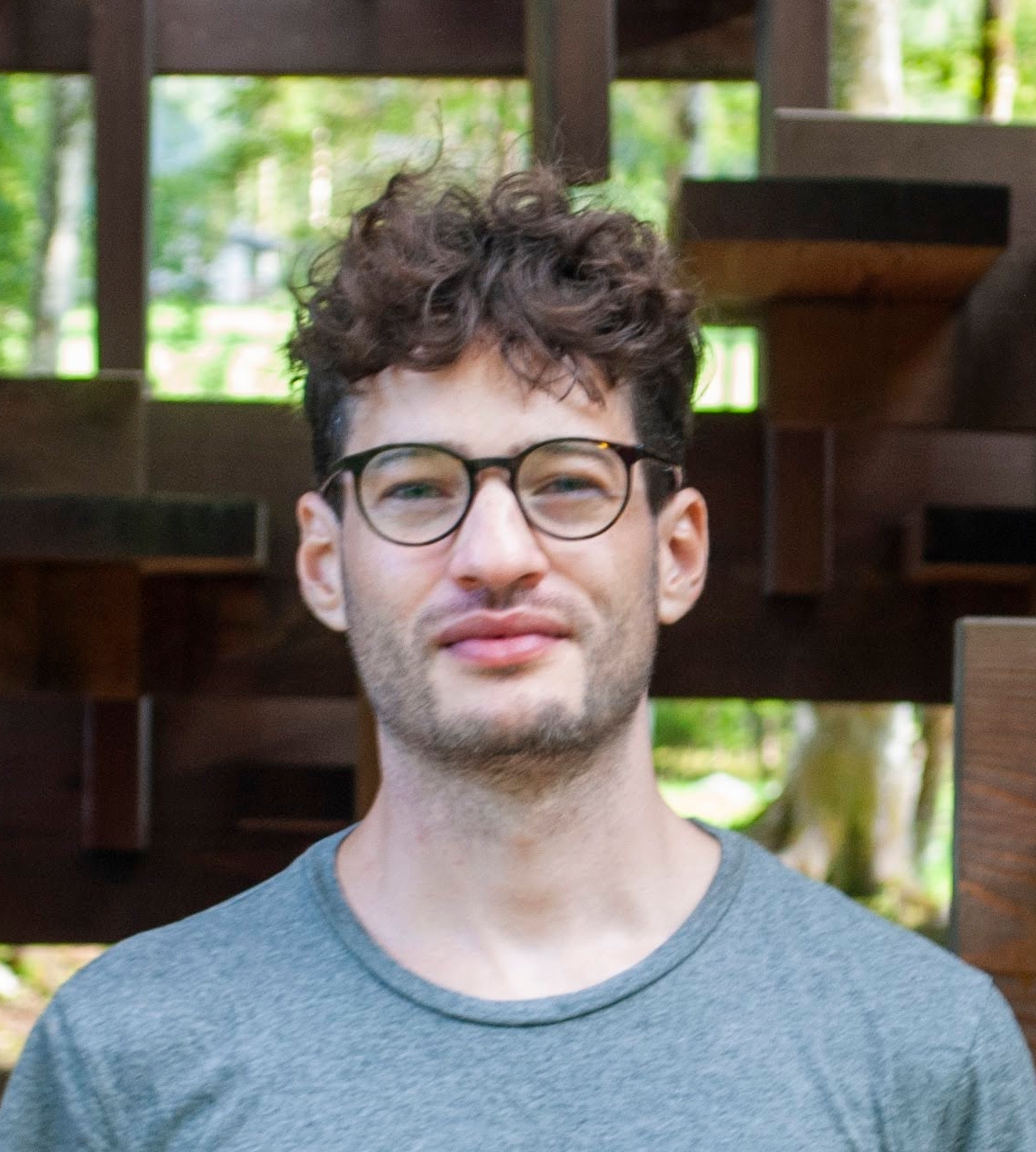}}]{Leandro de Souza Rosa}
Leandro de Souza Rosa has a BSc (2013) in Computer Engineering and obtained the Ph.D. in 2019 by the Institute of Mathematics and Computer Sciences at The University of Sao Paulo (ICMC-USP), Brazil, which was partially developed at the Department of Electrical and Electronic Engineering of Imperial College London. His interest areas are hardware architecture design, high-level synthesis and compilation tools, and code optimisation. He is currently working as a Post-Doc researcher at the Istituto Italiano di Tecnologia (IIT), Genova, Italy, focusing event-driven sensors for robotics. He has also served as an ad-hoc reviewer of the journals IEEE-TIM, IEEE-TCADICS, Int. J. of Embedded Systems (IJES),  Science China Information Sciences (SCIS),  Elsevier MICPRO, and IEEE Sensors Journal.
\end{IEEEbiography}

\begin{IEEEbiography}[{\includegraphics[width=1in,height=1.25in,clip,keepaspectratio]{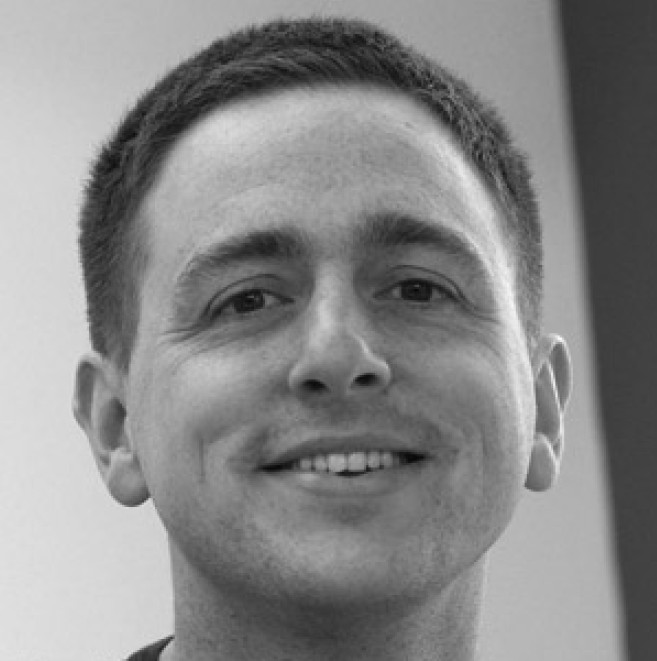}}]{Simeon Bamford}
Simeon Bamford completed his PhD in Neuroinformatics at the University of Edinburgh in 2009. Since then he has worked both academically and commercially on neural and neuromorphic engineering, in hardware and software, including event-based vision.
\end{IEEEbiography}

\begin{IEEEbiography}[{\includegraphics[width=1in,height=1.25in,clip,keepaspectratio]{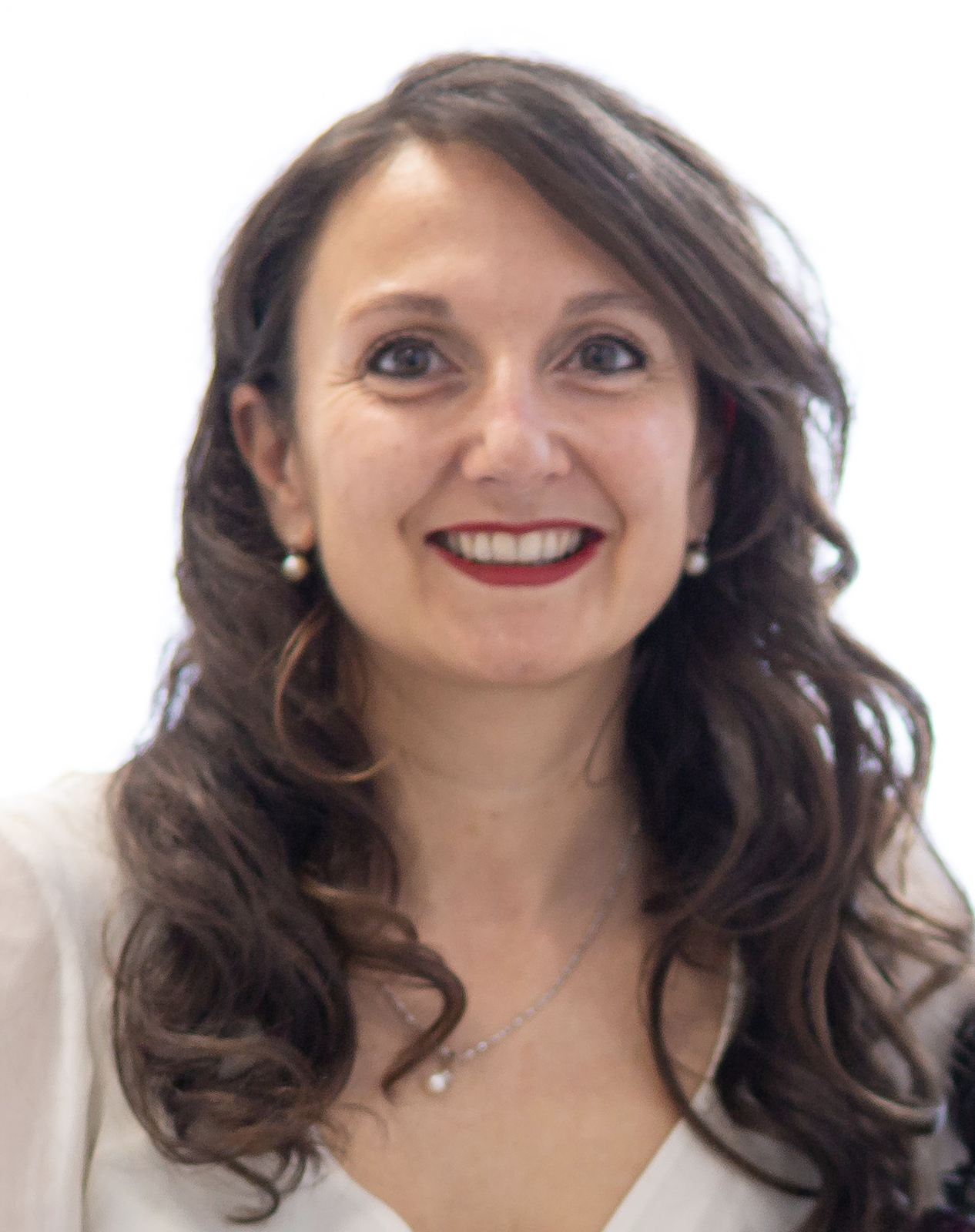}}]{Chiara Bartolozzi}
Chiara Bartolozzi (IEEE Member) is researcher at the Istituto Italiano di Tecnologia. She earned a degree in Engineering (with honors) at University of Genova (Italy) and a Ph.D. in Neuroinformatics at ETH Zurich, developing analog subthreshold circuits for emulating biophysical neuronal properties onto silicon and modelling selective attention on hierarchical multi-chip systems. She is currently leading the Event Driven Perception for Robotics group (www.edpr.iit.it), mainly working on the application of the "neuromorphic" engineering approach to the design of sensors and algorithms for robotic perception. She served as chair of the Neuromorphic Systems and Application Technical Committee of IEEE CAS and as co-general chair of IEEE Int. Conf. on AI Circuits and Systems AICAS2020.
\end{IEEEbiography}




\end{document}